\documentclass{article}

\usepackage{microtype}
\usepackage{graphicx}
\usepackage{subcaption}
\usepackage{booktabs} 

\usepackage{hyperref}
\usepackage{booktabs} 
\usepackage{siunitx}  
\usepackage{multirow}
\usepackage{tcolorbox}
\tcbuselibrary{breakable, skins}

\usepackage[preprint]{icml2026}

\usepackage{amsmath}
\usepackage{amssymb}
\usepackage{mathtools}
\usepackage{amsthm}

\usepackage[capitalize,noabbrev]{cleveref}

\theoremstyle{plain}

\theoremstyle{definition}

\theoremstyle{remark}

\usepackage[textsize=tiny]{todonotes}

\icmltitlerunning{Submission and Formatting Instructions for ICML 2026}

\begin{document}

\twocolumn[
  \icmltitle{DecisionLLM: Large Language Models for Long Sequence Decision Exploration}

  \begin{icmlauthorlist}
    \icmlauthor{Xiaowei Lv}{yyy,intern}
    \icmlauthor{Zhiling Zhang}{comp}
    \icmlauthor{Yijun Li}{comp}
    \icmlauthor{Yusen Huo}{comp}
    \icmlauthor{Siyuan Ju}{comp}
    \icmlauthor{Xuyan Li}{comp}
    \icmlauthor{Chunxiang Hong}{comp}
    \icmlauthor{Tianyu Wang}{comp}
    \icmlauthor{Yongcai Wang}{yyy}
    \icmlauthor{Peng Sun}{comp}
    \icmlauthor{Chuan Yu}{comp}
    \icmlauthor{Jian Xu}{comp}
    \icmlauthor{Bo Zheng}{comp}
  \end{icmlauthorlist}

  \icmlaffiliation{yyy}{Renmin University of China, Beijing China}
  \icmlaffiliation{comp}{Alibaba Group, Beijing, China}
  \icmlaffiliation{intern}{Work is done during the internship at Alibaba Group.}
  
  \icmlcorrespondingauthor{Yongcai Wang}{ycw@ruc.edu.cn}
  \icmlcorrespondingauthor{Bo Zheng}{bozheng@alibaba-inc.com}
  
  \icmlkeywords{Machine Learning, ICML}

  \vskip 0.3in
]



\printAffiliationsAndNotice{}  

\begin{abstract}
  Long-sequence decision-making, which is usually addressed through reinforcement learning (RL), is a critical component for optimizing strategic operations in dynamic environments, such as real-time bidding in computational advertising. The Decision Transformer (DT) introduced a powerful paradigm by framing RL as an autoregressive sequence modeling problem. Concurrently, Large Language Models (LLMs) have demonstrated remarkable success in complex reasoning and planning tasks.
  This inspires us whether LLMs, which share the same Transformer foundation, but operate at a much larger scale, can unlock new levels of performance in long-horizon sequential decision-making problem. 
  This work investigates the application of LLMs to offline decision making tasks. A fundamental challenge in this domain is the LLMs' inherent inability to interpret continuous values, as they lack a native understanding of numerical magnitude and order when values are represented as text strings. To address this, we propose treating trajectories as a distinct modality. By learning to align trajectory data with natural language task descriptions, our model can autoregressively predict future decisions within a cohesive framework we term DecisionLLM.
  We establish a set of scaling laws governing this paradigm, demonstrating that performance hinges on three factors: model scale, data volume, and data quality. 
  In offline experimental benchmarks and bidding scenarios, DecisionLLM achieves strong performance. Specifically, DecisionLLM-3B outperforms the traditional Decision Transformer (DT) by 69.4 on Maze2D umaze-v1 and by 0.085 on AuctionNet.
  It extends the AIGB paradigm and points to promising directions for future exploration in online bidding.
\end{abstract}

\begin{figure*}[h]
  \centering
  \includegraphics[width=0.9\linewidth]{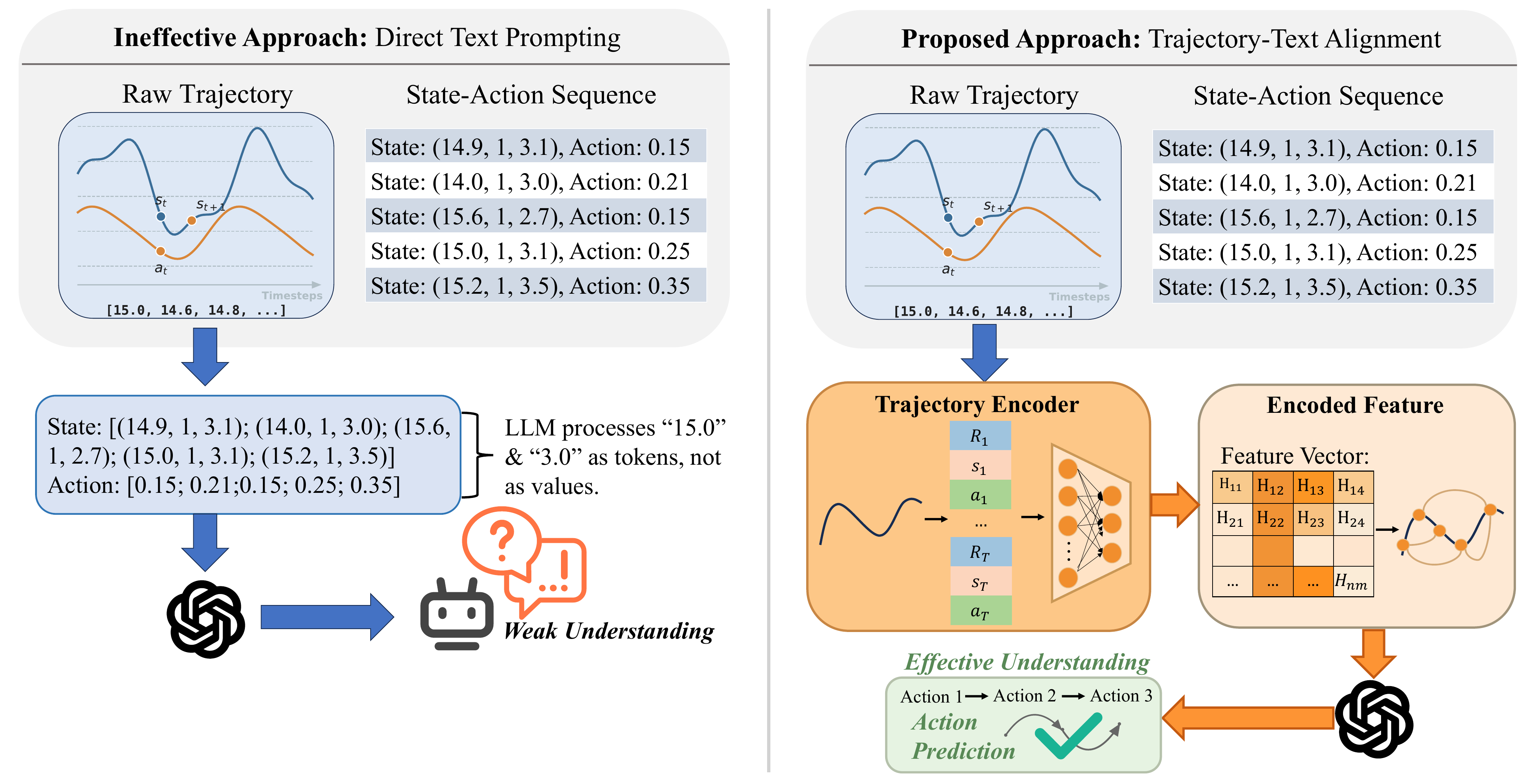}
  \caption{A comparison diagram of the prompt-based decision paradigm and the trajectory-text modality alignment-based paradigm.}
  \label{fig_fig1_sketch_map}
\end{figure*}

\section{Introduction}
 Addressing the challenge of long-sequence decision-making, where an agent must make coherent decisions over protracted time steps to achieve a long-term objective, is a cornerstone of traditional reinforcement learning. 
 Historically, reinforcement learning (RL) has long been the dominant approach to long-sequence decision-making 
 \cite{watkins1992q, wang2016dueling, schulman2017proximal,lillicrap2015continuous}. A critical context for addressing these problems is the offline setting, formally known as Offline Reinforcement Learning. \cite{kostrikov2021offline, kumar2020conservative,nair2020awac}. In this major domain, an agent learns entirely from a static, precollected dataset. Recently, this field has been invigorated by generative approaches, particularly those based on Diffusion \cite{rombach2022high, chi2023diffusion, wang2022diffusion} and Transformer \cite{vaswani2017attention, chen2021decision} architectures. These methods reframe decision-making as a sequence generation task, predicting future actions based on past sequences. However, we contend that the upper limits of their performance are far from being realized \cite{tarasov2023corl}. Meanwhile, Large Language Models (LLMs) have emerged as exceptionally potent sequence models, achieving significant success in complex domains such as autonomous driving \cite{li2025recogdrive, cui2023drivellm} and robotics \cite{kim2024openvla, intelligence2504pi0}. 
 LLMs demonstrate a significant capacity form zero-shot generalization, which allows its decision-making capability extends beyond mere imitation learning.
 Therefore, whether LLMs integration can unlock new levels of performance in long-sequence decision-making is a critical and open question right now.

Although LLMs have demonstrated great potential compared with traditional approaches, applying it to sequential decision-making still requires overcoming many challenges.
In decision-making tasks, continuous data is used, which has a modality gap with the LLMs' text-centric nature. 
Serializing the continuous trajectory data into a raw text prompt is the simplest approach, but this strategy is inherently flawed in practice. 
The reason is that LLMs are not natively sensitive to the quantitative meaning of numbers \cite{dziri2023faith}; as shown in Figure \ref{fig_fig1_sketch_map}, LLMs process ``3.0" and ``15.0" as specific tokens rather than as values with distinct magnitudes. For example, in prevalent environments like Maze2D \cite{fu2020d4rl}, where trajectory data are continuous variables, like states and actions. This limitation becomes critical; 
LLMs struggle to capture how a current action shapes future numerical trajectories accurately, which means relying solely on a text-only representation is neither sufficient nor accurate.
Given this, a central challenge for harnessing LLMs in long-sequence decision-making is to achieve effective alignment between input and output trajectories in a way that preserves the quantitative meaning of numbers. 
Drawing inspiration from the paradigm of large multimodal models \cite{liu2024improved, xie2024chatts}, we introduce a trajectory-text alignment mechanism that treats trajectories as a distinct data modality to bridge the gap between text and continuous sequence data.

\begin{figure*}[h]
  \centering
  \includegraphics[width=0.9\linewidth]{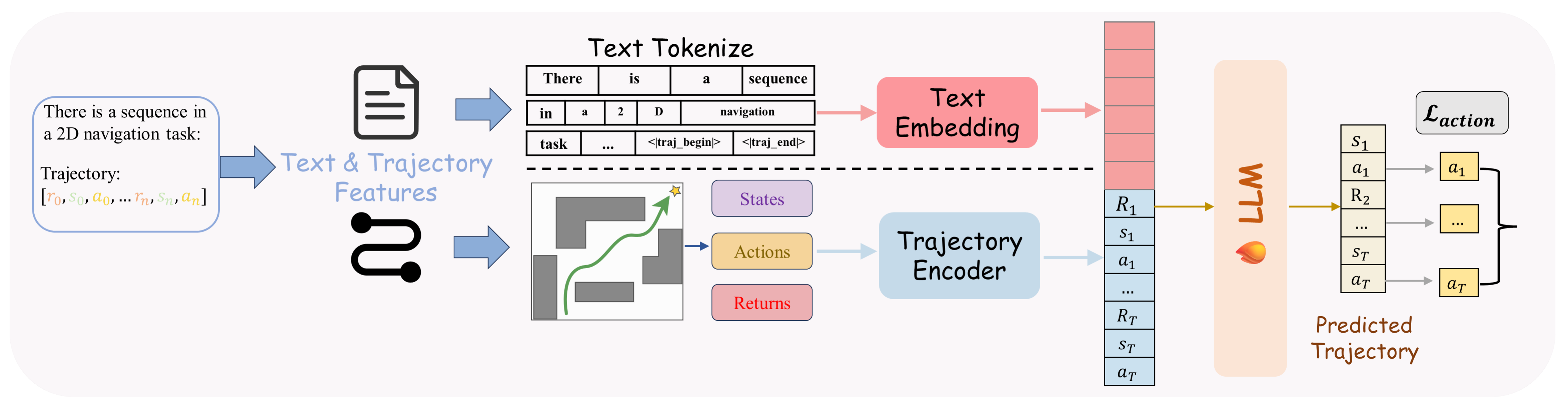}
  \caption{Overview diagram of the framework.}
  \label{fig_overview}
\end{figure*}


In this work, we present DecisionLLM, a multimodal framework that, to our knowledge, is the first to jointly process text and trajectories for long sequence decision making by treating trajectories as a distinct modality. DecisionLLM uniquely fuses textual instructions with encoded trajectory features to autoregressively generate decisions. Our architecture employs two critical components to interface between the trajectory modality and the text modality.
Firstly, We employ a trajectory encoder which processes the input sequence of states, actions, and returns-to-go into a compact embedding used for alignment; then, we concatenate it with the text's embedding; finally, a linear projection head is tasked with mapping the LLM's final contextualized embedding back into the continuous action space to generate the ultimate action prediction.
Given that the standard pre-training of LLMs does not encompass an understanding of trajectory data, we finetune our model using an autoregressive objective to predict the action on the current timestep, conditioned on the historical trajectory. The ground-truth actions from the offline trajectories serve as the training labels for this task. In practice we have observed that clarifying data quality standards can enhance practical outcomes, including filtering out low-quality trajectories and reducing the weight of low-reward steps.
Furthermore, in the bidding scenario, DecisionLLM broadens the paradigm of traditional AI-Generated Bidding (AIGB) \cite{guo2024generative}, our experiments on the AuctionNet \cite{su2024auctionnet} benchmark confirm the effectiveness of this extended approach.
Meanwhile, our empirical results reveal clear scaling laws governing this paradigm: performance systematically improves with increases in model parameters count, data volume, and data quality.

Our contributions are summarized as follows:

\textbf{New Modality and Architecture.} We introduce a novel approach that treats trajectories as a distinct data modality. And we propose DecisionLLM, a multimodal architecture designed to predict future actions based on textual task descriptions and historical trajectories. 

\textbf{Scaling Laws.} Our validation of these scaling laws reveals a crucial insight: model and data scaling are not independent factors but are synergistically linked. 

\textbf{Data Quality.} We also underscores the critical role of data quality in the performance of DecisionLLM. Model capabilities can be enhanced by filtering low-quality samples.

\textbf{Experimental Performance.} The efficacy of our approach is validated through experiments on maze2d-umaze-v1 and AuctionNet benchmark. DecisionLLM-3B significantly outperforms the Decision Transformer, achieving performance gains of 69.4 points and 0.058 score respectively. 

\begin{figure}[h]
    \centering 

    \begin{subfigure}[b]{0.22\textwidth}
        \centering
        \includegraphics[width=\linewidth]{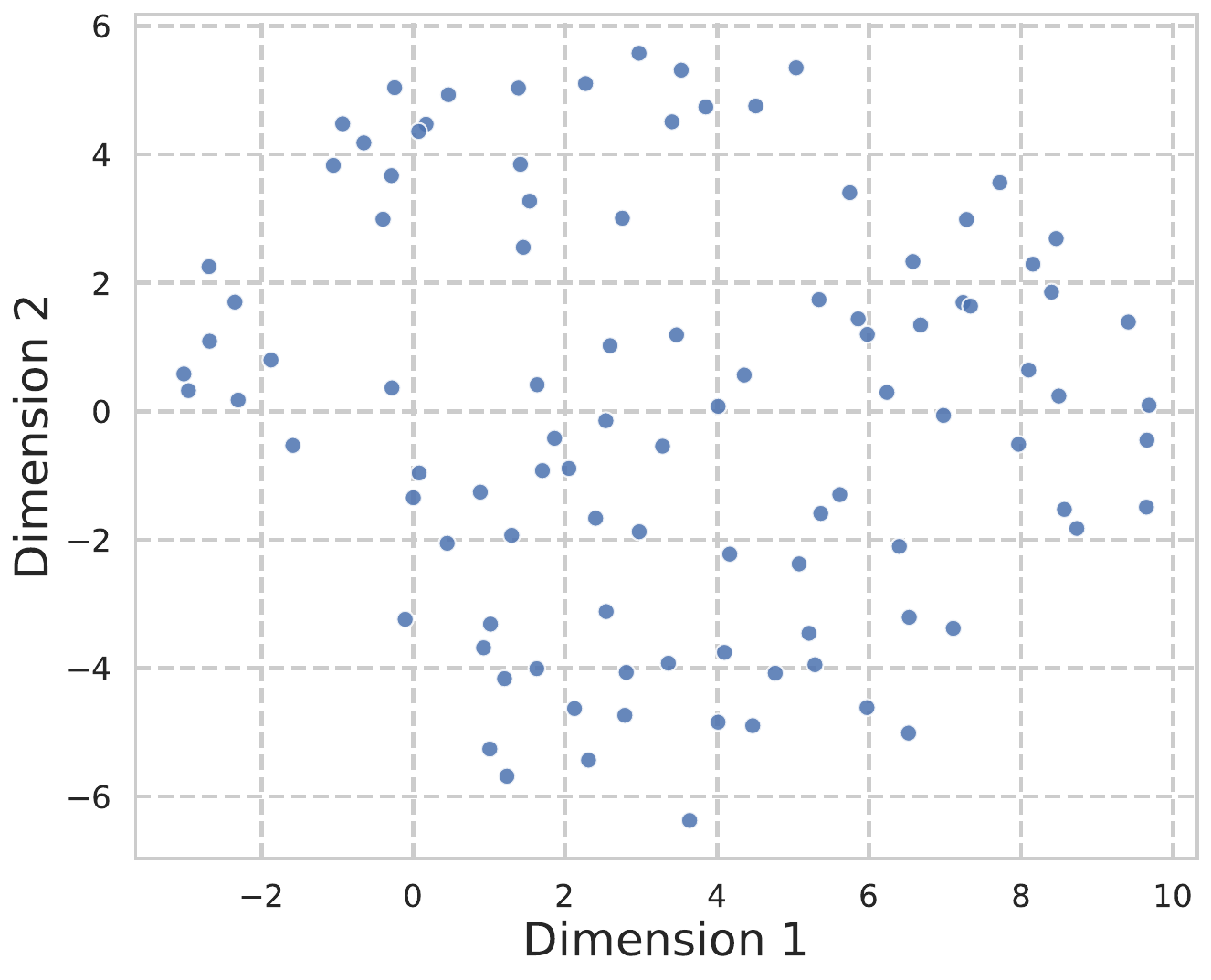}
        \caption{t-SNE of LLM (prompt)}
        \label{fig:tsne_llm_prompt}
    \end{subfigure}
    \hfill 
    \begin{subfigure}[b]{0.22\textwidth}
        \centering
        \includegraphics[width=\linewidth]{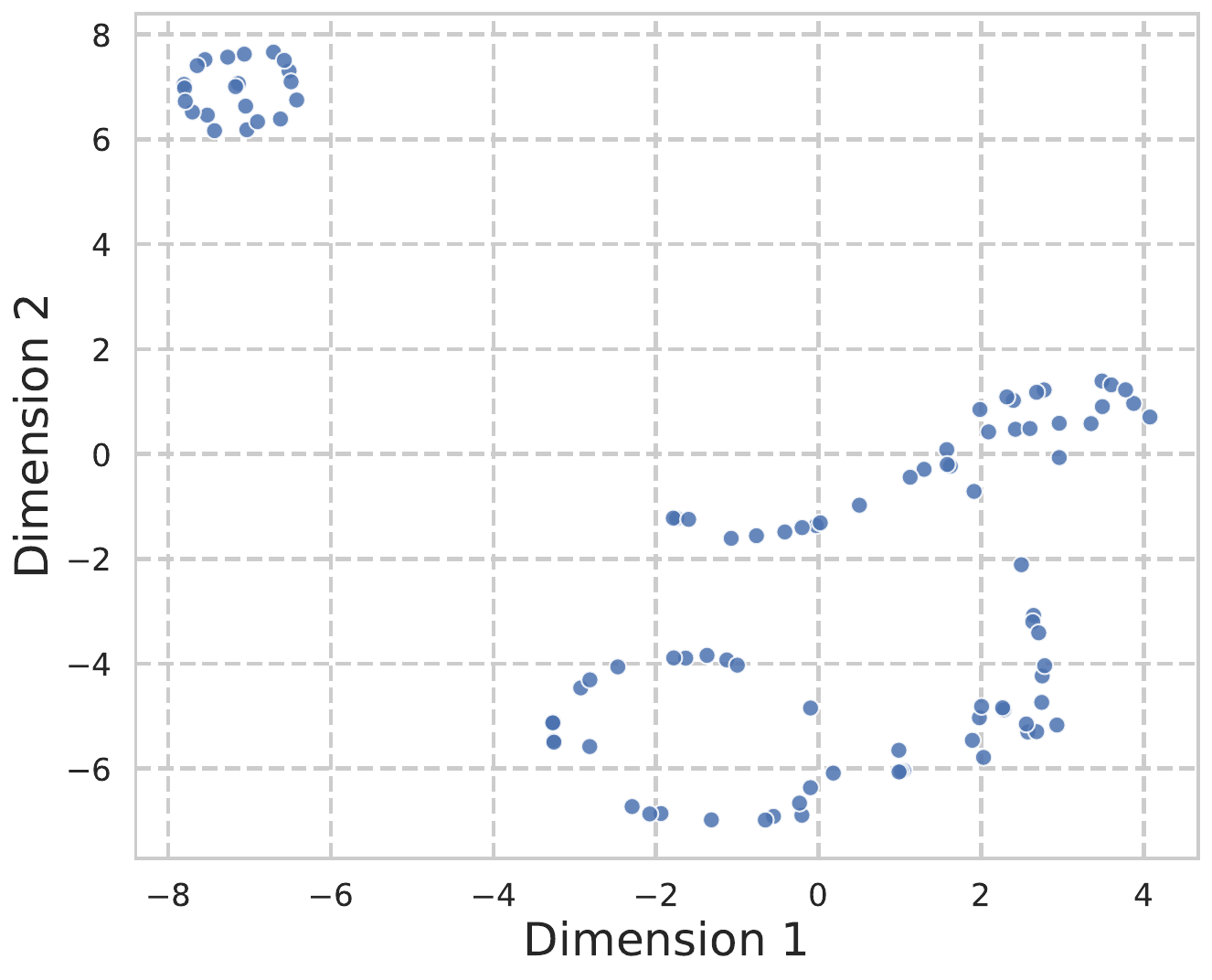}
        \caption{t-SNE of DecisionLLM}
        \label{fig:tsne_decisionllm}
    \end{subfigure}

    \par\vspace{1em} 

    \begin{subfigure}[b]{0.22\textwidth}
        \centering
        \includegraphics[width=\linewidth]{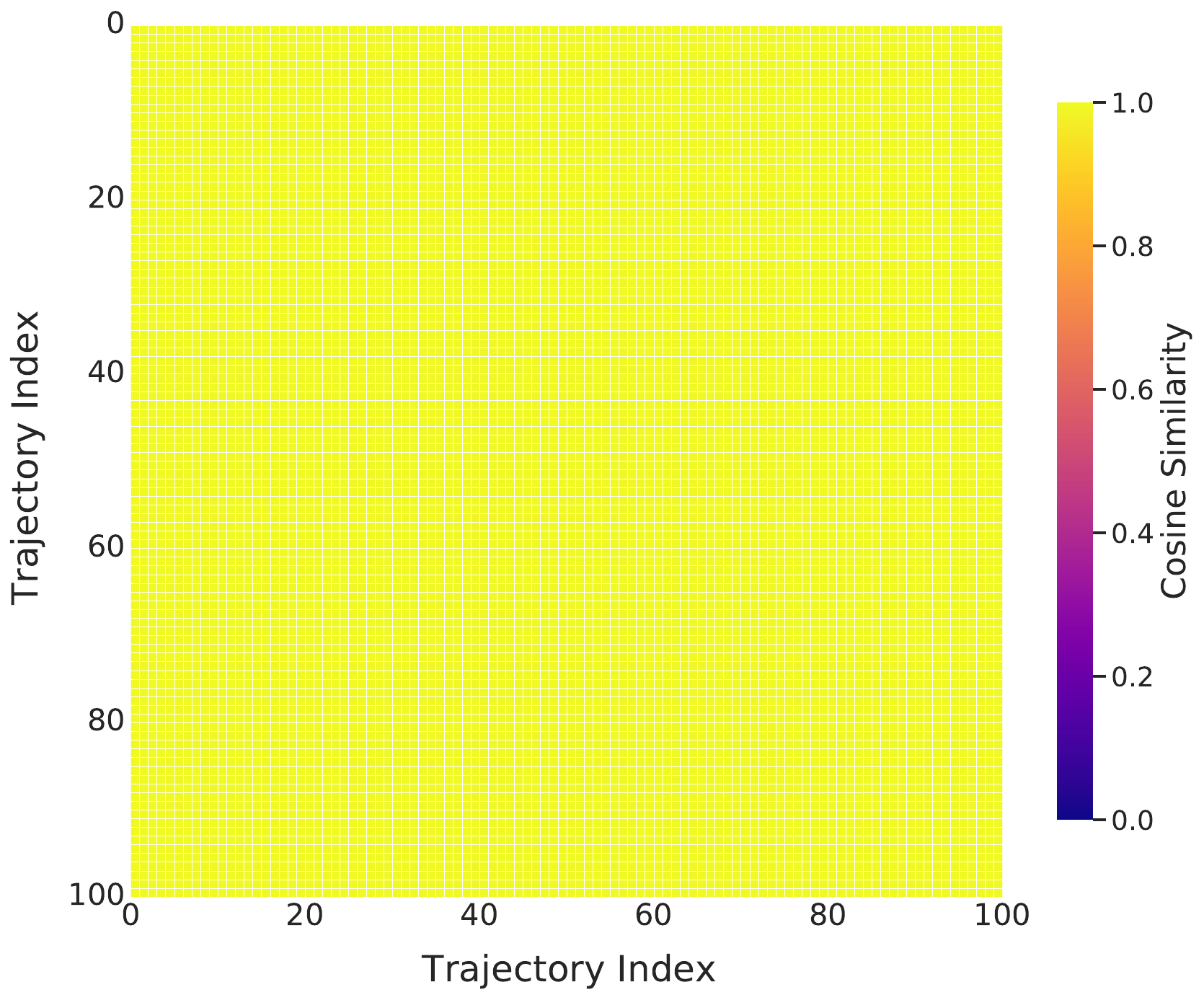}
        \caption{heatmap of LLM (prompt)}
        \label{fig:heatmap_llm_prompt}
    \end{subfigure}
    \hfill
    \begin{subfigure}[b]{0.22\textwidth}
        \centering
        \includegraphics[width=\linewidth]{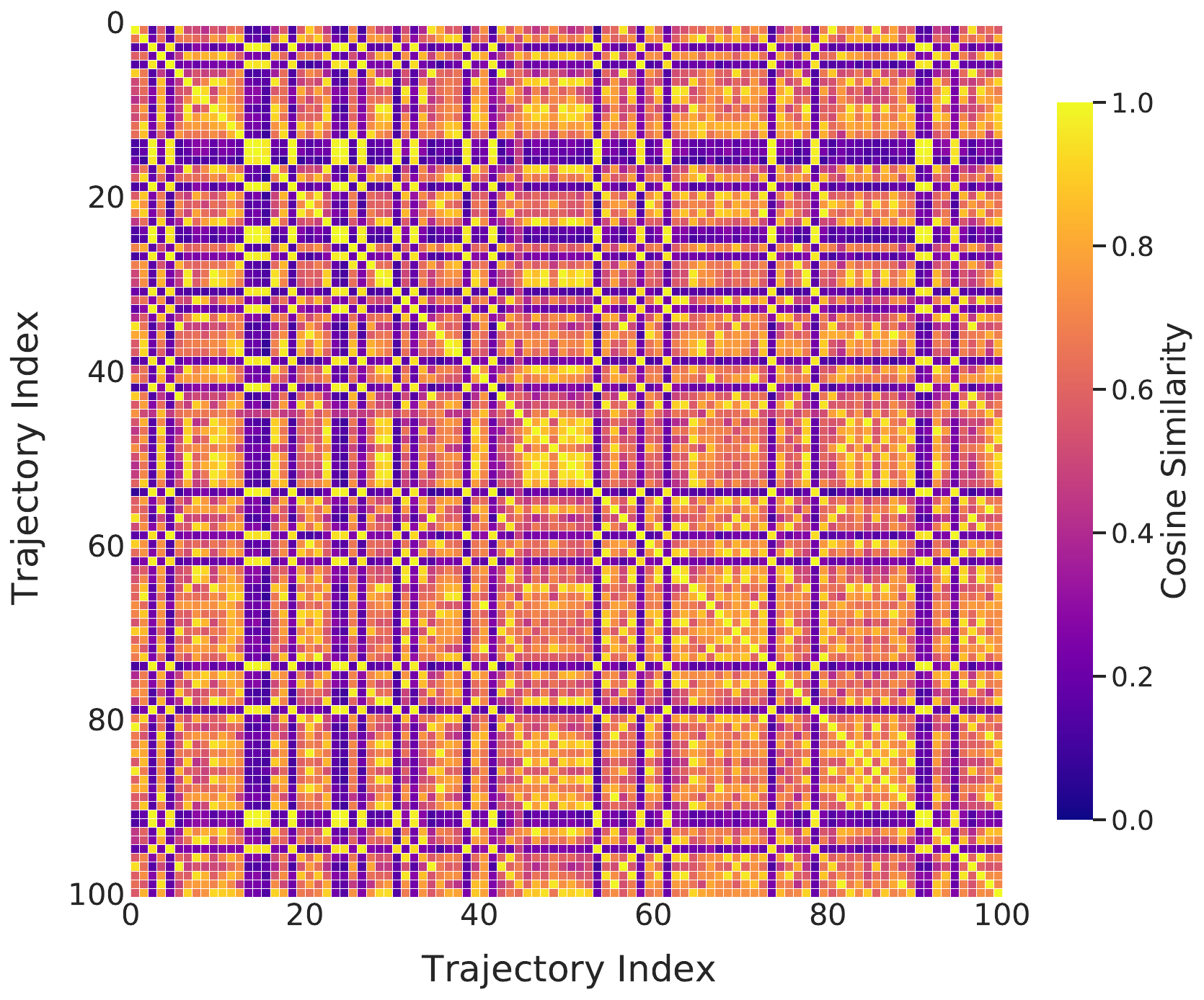}
        \caption{heatmap of DecisionLLM}
        \label{fig:heatmap_decisionllm}
    \end{subfigure}

    \par\vspace{1em}

    \begin{subfigure}[b]{0.4\textwidth}
        \centering
        \includegraphics[width=\linewidth]{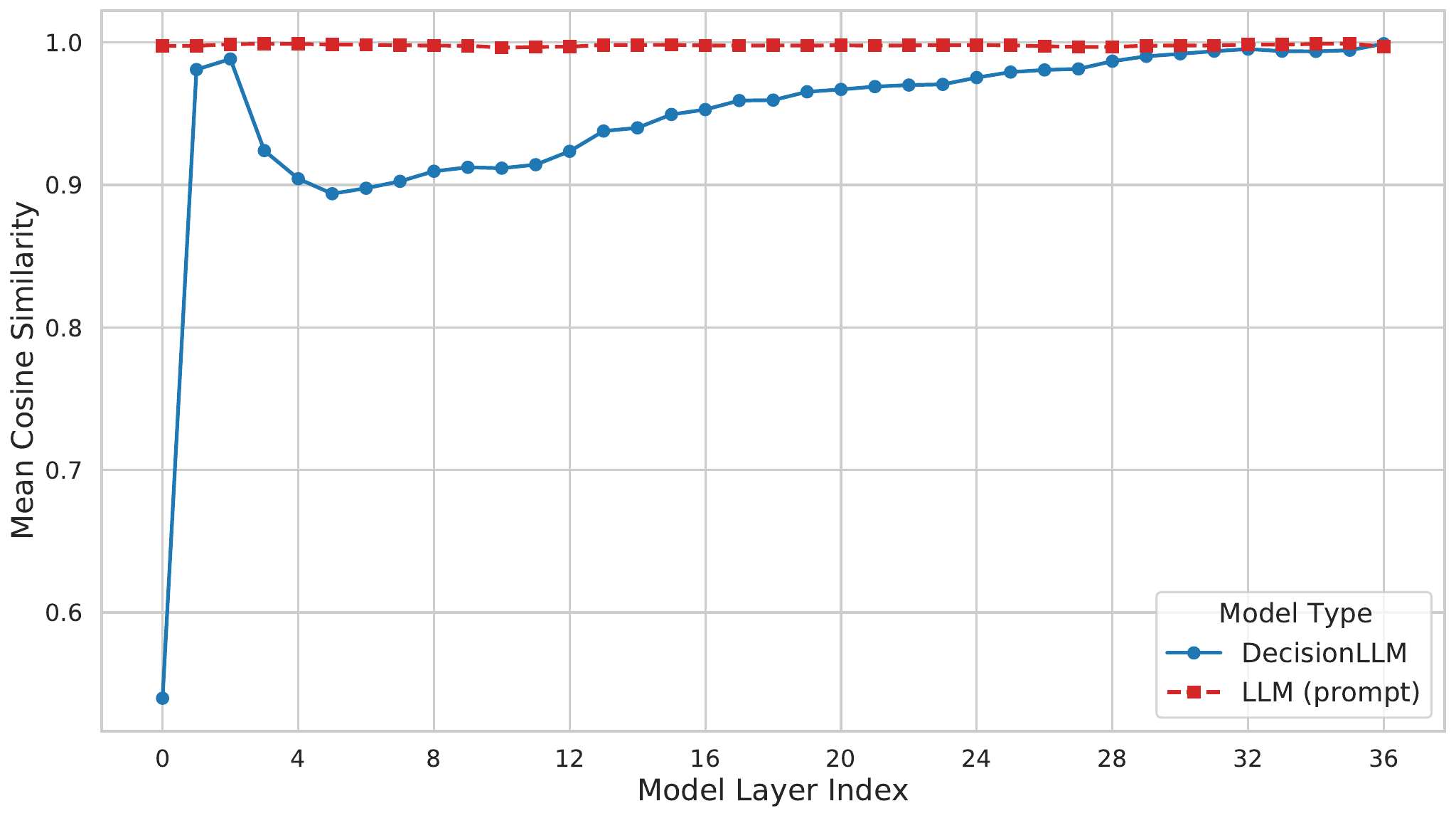}
        \caption{Mean cosine similarity comparison}
        \label{fig:mean_cosine_similarity}
    \end{subfigure}

    \caption{Embedding analysis graph based on prompt-based trajectory and trajectory-modal (DecisionLLM) input. }
    \label{fig:explain_trajectory}
\end{figure}

\section{Analysis}

To investigate the representational differences between treating trajectories as text prompts versus a distinct modality, we randomly sampled 100 trajectories and extracted their input layer embeddings using both the prompt-based LLM and the trajectory-modal Decision LLM. We applied t-SNE to project these high-dimensional embeddings into a two-dimensional space for visualization, as shown in Figures \ref{fig:tsne_llm_prompt} and \ref{fig:tsne_decisionllm}. Furthermore, we computed the pairwise cosine similarities to generate similarity heatmaps (Figures \ref{fig:heatmap_llm_prompt} and \ref{fig:heatmap_decisionllm}). Finally, to quantify the representational evolution, we calculated and plotted the average cosine similarity across all model layers, as illustrated in Figure \ref{fig:mean_cosine_similarity}.

The experimental results yield the following insights:

\textbf{1.Representation collapse in prompt-based encoding.} 
    Trajectories processed as text prompts struggle to achieve separability. The $t$-SNE visualization reveals that embeddings from the prompt-based LLM are unstructured and difficult to classify, exhibiting a lack of distinct clustering. The corresponding heatmaps show uniformly high cosine similarity across pairwise trajectories. This phenomenon indicates that the model's attention is dominated by the static textual templates of the prompt rather than the dynamic numerical variances, resulting in a failure to capture the fine-grained physical characteristics of the trajectories.

\textbf{2.Effective modal representation by DecisionLLM.} 
    In contrast, DecisionLLM successfully treats trajectories as a distinct modality. The $t$-SNE visualization demonstrates clear separation between clusters, highlighting the model's ability to effectively capture and disentangle trajectory features. Furthermore, the heatmap reveals distinct pairwise differences, confirming that the model preserves data diversity and discriminative physical features. This validates the practical significance of encoding trajectories as a dedicated modality rather than raw text.

\textbf{3. Hierarchical abstraction from physical signals to semantic intents.} 
    While the prompt-based LLM maintains a constantly high mean cosine similarity, DecisionLLM exhibits a clear evolutionary trend: similarity is low in the initial layers but converges to near 1.0 in the final layers. This signifies a healthy abstraction process where low-level physical signals are highly discriminative at the input stage, and are gradually transformed into unified high-level semantic intents for final decision-making.

\section{Methodology}

\subsection{Overview}
Our work introduces a paradigm shift for direct decision-making with LLMs. We tackle the model's numerical insensitivity by treating trajectories as a first-class, non-textual modality. By co-training on aligned trajectory and task description within an autoregressive framework, we empower the LLM to ground its reasoning in offline long sequential data and generate effective actions. The power of this paradigm is twofold: first, it harnesses the vast, generalizable knowledge embedded in large-scale pre-trained models; second, it leverages explicit task descriptions via the text modality to contextualize the decision-making process. We instantiate this paradigm in our proposed architecture, DecisionLLM (Figure \ref{fig_overview}). By framing decision-making as a LLM task, we directly inherit its well-established scaling properties. Our subsequent analysis is therefore dedicated to empirically verifying these scaling laws with respect to data volume, parameter count, and the crucial role of data quality, achieved through targeted filtering. In subsequent subsections, we will detail the model design, training, and inference processes, as well as the corresponding data augmentation methods.

\subsection{Trajectory Modal Embedding}
\begin{figure}[h]
  \centering
  \includegraphics[width=\linewidth]{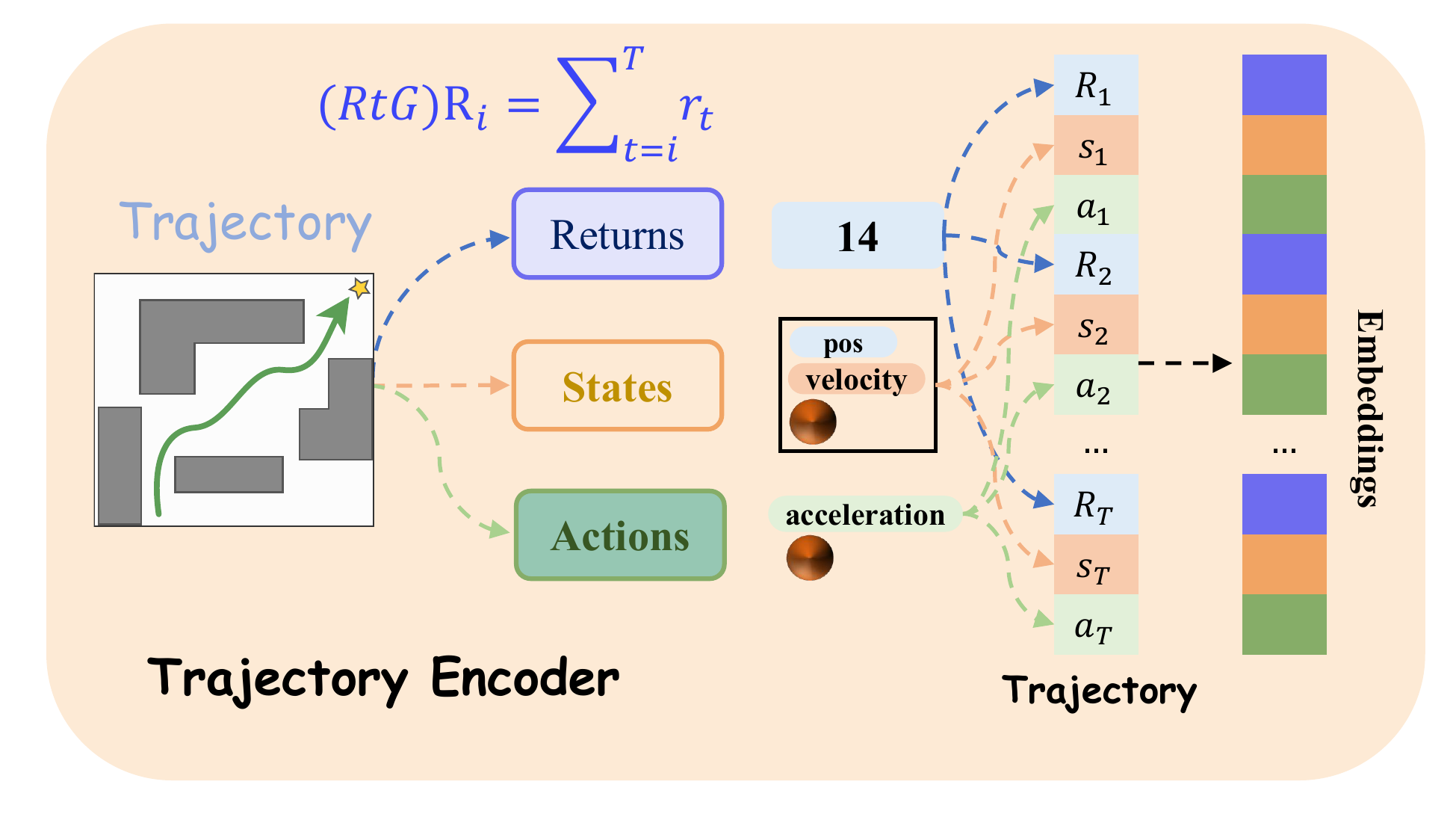}
  \caption{The specific architecture of the trajectory encoder.}
  \label{fig_trajectory_encoder}
\end{figure}


In Offline Reinforcement Learning, long-sequence decision making can be formulated as a Markov Decision Process (MDP), specified by $(\mathcal{S}, \mathcal{A}, P, \mathcal{R})$. The MDP tuple consists of states $s \in \mathcal{S}$, actions $a \in \mathcal{A}$, transition dynamics $P(s^\prime\mid s, a)$, and a reward function $r = \mathcal{R}(s,a)$. We use $s_t$, $a_t$, and $r_t = \mathcal{R}(s_t, a_t)$ to denote the state, action and reward at timestep $t$, respectively.
A trajectory $\tau = (s_t, a_t, r_t)_{t=0}^T$ is a sequence of states, actions, and rewards. 
The goal is to learn a policy $\pi(a|s)$ that maximizes the expected return, where the return-to-gos (Rtgs or Returns) is defined as $\hat{R_t} = \sum_{t'=t}^{T} r_{t'}$. 
In the offline setting, learning is constrained to a static dataset of trajectories, precluding further environmental interaction and making the learning problem susceptible to distributional shift.

Following the architectural paradigm of the Decision Transformer (DT) \cite{chen2021decision}, we first encode the three core components of a trajectory: Rtgs, states and actions, using distinct embedding layers. These modality-specific embeddings are then interleaved according to their timestep to form a single, unified sequence, the sequence will be served as the input to the transformer model. In the result, each trajectory is mapped into a sequence of interleaved triplets, forming the input sequence:
\begin{equation}
  \tau = ( \hat{R_1}, s_1, a_1, \hat{R_2}, s_2, a_2, \dots, \hat{R_T}, s_T, a_T)
\end{equation}

Following this, we transform the raw returns, states, and actions into dense vector embeddings. 
Specifically, each component (Rtgs, State, Action) is independently projected into a high dimensional space using a dedicated linear layer. 
These individual component embeddings are then concatenated to form the trajectory's input embedding. 
To incorporate temporal information, we add a positional encoding to each timestep in the sequence. 
The final trajectory embedding $\tau_e$ is constructed by interleaving the feature vectors of returns, states, and actions across time steps, thereby preserving the original sequential characteristic information. This directly circumvents the well-known issue of LLMs' poor numerical sensitivity. Consequently, we propose treating the entire trajectory as a distinct, non-textual modality, processing it using an architecture analogous to that of Multimodal Large Language Models.

\subsection{DecisionLLM Model Structure}

As illustrated in Figure \ref{fig_overview}, our model architecture is designed to process textual task descriptions and raw trajectory sequences. The text input is first tokenized and then converted into vector embeddings using the LLM's native embedding layer. 
Concurrently, the trajectory sequence is processed by the trajectory encoder, as detailed in Section 3.2, to produce a comprehensive embedding.

To fuse these two modalities, we introduce a novel prompting strategy using special placeholder tokens, \texttt{<|traj\_begin|>} and \texttt{<|traj\_end|>}. These tokens are inserted into the textual prompt to designate a slot for the trajectory information. During processing, the computed trajectory embedding effectively substitutes the embeddings of these placeholders, thereby injecting the entire trajectory context into the LLM's input sequence. For the output stage, the model employs an autoregressive decoding process to predict subsequent actions. 
Then, we add an additional action head and use a linear head to map the output logits to the action space. 
Finally, an action mapping layer is applied to the transformer's output logits to generate the predicted action, $\hat{A}$. The training loss $L$ is a Mean Square Error (MSE) function of the predicted action $\hat{A}$ and the actual action $A$.

A key architectural innovation of DecisionLLM is its handling of trajectory data as a non-textual modality. This approach directly circumvents the LLM's fundamental limitation in processing numerical data encoded as text. As a result, the model builds a native and effective representation of trajectory sequences, enabling a robust mapping from input history to output actions.

\subsection{DecisionLLM Training and Inference}

\subsubsection{DecisionLLM Training}

Given that the training process employs an autoregressive approach, the training data is derived exclusively from the trajectory itself, which is based on offline sampled trajectory dataset. Specifically, the complete trajectory serves as the input to the model, while its shifted version is used as the training labels. The task description is derived from the environment's basic information and includes the task objectives, state space, action space, design of the reward function, and other relevant details. Since the LLM inherently lacks understanding of trajectory modalities, it is necessary to train both its input and output components to correctly interpret such data and generate accurate action predictions. Accordingly, we optimize full parameters of DecisionLLM, including the linear layers in both the input and output modules, as well as all parameters within the LLM itself.

In addition, since the actual sampled trajectories in some scenarios can be excessively long, a sliding window approach is employed during training. This restricts the model input to trajectories from the most recent $t$ time steps, thereby avoiding issues associated with processing very long sequences. We will introduce more parameter settings and details about training in Section 4.

\subsubsection{DecisionLLM Inference}
Once trained, the model generates actions autoregressively based on its history. Initially, we provide the model with a target return $\hat{R_1}$ and the initial state $s_1$, conditioning it to predict the first action, $a_1$. 
This action is then executed in the environment, yielding the next state $s_2$ and a reward $r_1$. 
The target Rtgs is subsequently updated (e.g., $R_2 = R_1 - r_1$). 
This cycle is repeated: the newly formed sequence, incorporating the updated return $R_2$ and state $s_2$, is used to predict the next action, $a_2$. 
This interactive loop continues until the episode terminates or a predefined maximum length is reached. 


\subsection{Data Quality Improvement}
The training paradigm of DecisionLLM is fundamentally a form of imitation learning. 
Its objective is to distill effective policies from an offline dataset of historical trajectories. 
The model learns to associate high-return sequences with specific actions, thereby enabling it to generalize these successful behaviors to similar, unseen scenarios. 
Consequently, the performance of this imitation-based approach is critically sensitive to the quality and composition of the training data.

Given the substantial computational cost of training LLM and the sensitivity of our approach to data quality, a rigorous data filtering strategy is essential. We employ a return-based threshold to exclude low-quality trajectories, optimizing both data quality and training efficiency. Conversely, for suboptimal steps within valid trajectories, rigid filtering risks hindering exploration. To address this, we adopt a reweighting strategy that attenuates the influence of low-quality exploration without sacrificing the breadth of the state space coverage.

\section{Evaluation}
In this section, we conduct a evaluation of DecisionLLM and address the following research questions (RQs):

\textbf{RQ1.Performance}: How effectively does DecisionLLM perform in the target tasks?

\textbf{RQ2.Scaling Laws}: How do the scaling laws of DecisionLLM behave with respect to model parameter count and dataset size?

\textbf{RQ3.Data Quality}: To what extent does the quality of training data influence the performance of DecisionLLM?

\textbf{RQ4.Impact of Pretrained Parameters}: What effect does initializing with pretrained parameters have on the model’s downstream performance?


\subsection{Experimental Setup}
\subsubsection{Evaluation Tasks.} To comprehensively evaluate the model's performance, we employed the D4RL \cite{fu2020d4rl} open-source offline reinforcement learning benchmark, focusing on tasks with long sequence decisions, such as Maze2D. We strictly followed the benchmarking methodology established in CORL \cite{tarasov2023corl} to ensure consistent and fair comparisons. In cases where experiments involved dataset expansion or data quality filtering, this is explicitly indicated in the respective sections. The primary benchmark datasets used in our evaluation include:

\noindent\textbf{Maze2D.} A navigation environment in which the objective is to guide a ball to a target location as efficiently as possible. In maze tasks, rewards are only given when ball reach near the end point, and the intermediate steps have an impact on the final result, which is highly consistent with long-sequence decision-making tasks. The sparse reward structure makes this task particularly challenging. We mainly selected maze2d-umaze-v1 for the experiment.

\noindent\textbf{AuctionNet.} In addition, to verify the effectiveness of the paradigm in more scenarios, we selected a benchmark AuctionNet \cite{su2024auctionnet} for an automatic bidding scenario, which simulates the completeness and complexity of real advertising auctions, including the ad opportunity generation module, the bidding module, and the complex module. Therefore, the state and action space is more complex.



\subsubsection{Evaluation Metrics.} For maze2D task, two key metrics are employed to evaluate the experimental results: reward and normalized score provided by D4RL. To ensure statistical reliability, all results are averaged over 100 independent evaluation runs. For AuctionNet, we evaluated the results using the final scores from the online assessments. We used 48 players and 7 episodes to ensure consistency of the results. 

\subsubsection{Training Details.} Our models are all trained based on the pretrained parameters of the Qwen2.5-Instruct model at different scales. The batch size of training is set to 64, the learning rate is 1e-5, and the window size is set to 20. All experiments were conducted on a server equipped with 8 NVIDIA A100 (40G) GPUs. Our implementation leverages the llama-factory \cite{zheng2024llamafactory} training framework, with distributed training accelerated by DeepSpeed-ZeRO stage 2. Models were trained for a total of 5 epochs, using a cosine annealing learning rate schedule. During training, we performed evaluations every 200 steps. For each experimental run, we report the peak performance achieved across all evaluation checkpoints. More details can be found in Appendix \ref{exp_appendix}.

\begin{table}[htbp]
  \centering
  \caption{Comparison experiments on Maze2D-umaze-v1.}
  \label{tab:maze2d_comparison}
  \begin{tabular}{lcc}
    \toprule
    Model & Return & Score \\
    \midrule
    BC & $46.06 \pm 25.05$ & $16.09 \pm 0.87$ \\
    TD3+BC & $160.94 \pm 46.15$ & $99.33 \pm 16.16$ \\
    CQL & $150.89 \pm 42.70$ & $92.05 \pm 13.66$ \\
    IQL & $94.12 \pm 29.69 $ & $50.92 \pm 4.23$ \\
    DT & $111.94 \pm 47.79 $ & $63.83 \pm 17.35$ \\
    DT-extended & $175.5 \pm 65.27 $ & $109.9 \pm 47.29$ \\
    \midrule
    \multicolumn{3}{l}{\textit{(LLM-based methods)}} \\
    LLM-prompt & $13.17 \pm 35.12$ & $-7.74 \pm 25.45$ \\ 
    LLM-ht & $45.58\pm 105.42$ & $15.75 \pm 76.38$\\
    LLM-hpt & $45.82 \pm 36.16$ & $15.92 \pm 26.20$ \\
    DecisionLLM(0.5B) & $204.45 \pm 71.74$ & $130.86 \pm 51.98$ \\
    DecisionLLM(1.5B) & \textbf{224.19 $\pm$ 48.74} & \textbf{145.16 $\pm$ 35.32} \\
    DecisionLLM(3B) & $220.18 \pm 52.48$ & $142.26 \pm 38.02$ \\
    \bottomrule
  \end{tabular}
\end{table}

\begin{table}[htbp]
  \centering
  \caption{Comparison experiments on AuctionNet.}
  \label{tab:auctionnet_comparison}
  \begin{tabular}{lc}
    \toprule
    Model & Score \\
    \midrule
    BC & $0.385$ \\
    TD3+BC & $0.317$ \\
    CQL & $0.357$ \\
    IQL & $0.388$ \\
    DT & $0.313$ \\
    \textbf{DecisionLLM(3B)} & \textbf{0.398} \\
    \bottomrule
  \end{tabular}
\end{table}

\subsubsection{Baselines.}
We compare against several baseline methods, primarily from offline reinforcement learning. These include pure RL-based offline algorithms such as IQL \cite{kostrikov2021offline} and CQL \cite{kumar2020conservative}, as well as supervised learning-based approaches like Behavioral Cloning (BC), TD3+BC \cite{fujimoto2021minimalist}, Decision Transformer (DT) \cite{chen2021decision}, and DT-extended (i.e., the version that utilizes the same expanded dataset as DecisionLLM). We also evaluated the LLM-prompt (i.e., prompting the model in textual form to predict actions based on the current state), the LLM-hp (i.e., directly concatenating the trajectory information into the prompt as textual numerical strings), and LLM-hpt (i.e., the model trained under the LLM-hp paradigm). These models were all evaluated or trained using Qwen2.5-3B-Instrcut. Our proposed model, DecisionLLM, was implemented and evaluated at 0.5B, 1.5B, and 3B parameters.



\subsection{Performance}


Table \ref{tab:maze2d_comparison} and Table \ref{tab:auctionnet_comparison} provide a comprehensive comparison of our model's performance on the Maze2D-umaze-v1 and AuctionNet benchmark. For Maze2D-umaze-v1 task, these RL-based offline algorithms' scores are form CORL \cite{tarasov2023corl}. Compared to the DT, DecisionLLM achieves a best case performance improvement of 82 points; compared to other RL algorithms, DecisionLLM achieves a best case performance improvement of 45 points.

Due to computational resource constraints, we focused our evaluation of DecisionLLM (3B) on the AuctionNet benchmark. Despite these limitations, the model demonstrated superior performance compared to other RL baselines. Notably, to rigorously assess the model's capability on sparse data, we trained it directly on the limited benchmark dataset without applying any data augmentation.

Notably, while model performance continues to improve from DT to DecisionLLM (0.5B) to DecisionLLM (1.5B), it exhibits little decrement at DecisionLLM (3B). We will provide further analysis and discussion in Section 4.3. A key finding relates to the model's data efficiency. While we carefully selected a large, high-quality dataset to ensure robust training, our experiments show that DecisionLLM achieves state-of-the-art performance using only a small portion of the data. This improvement is achieved by sampling only a small portion of the windowed trajectories from a large sample of data. This demonstrates that the model is able to efficiently extract a strong learning signal from a relatively small number of high-quality demonstrations. More details can be found in Appendix \ref{performance_appendix}.

\subsection{Property Analysis}
To further validate a series of properties of DecisionLLM, we conducted additional analytical experiments, including related scaling laws, data quality analysis, and corresponding pretraining parameters. Due to resource constraints, the following experiments were performed only in the Maze2D.
\subsubsection{Scaling Laws}

\begin{figure}
    \centering
    \begin{subfigure}[b]{0.23\textwidth}
        \includegraphics[width=\textwidth]{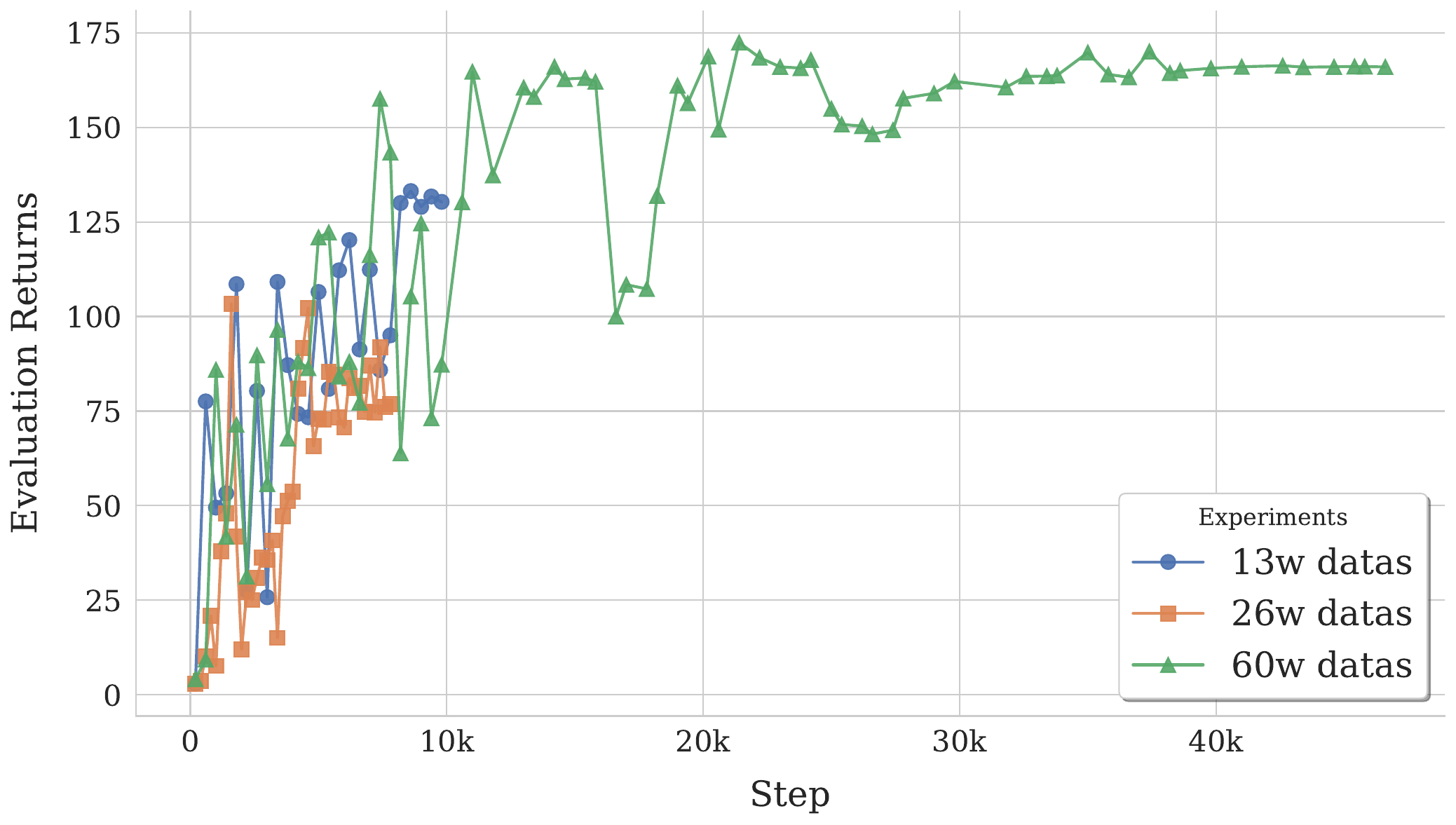}
        \caption{Returns 140}
        \label{fig:data_scaling_140}
    \end{subfigure}
    \hfill
    \begin{subfigure}[b]{0.23\textwidth}
        \includegraphics[width=\textwidth]{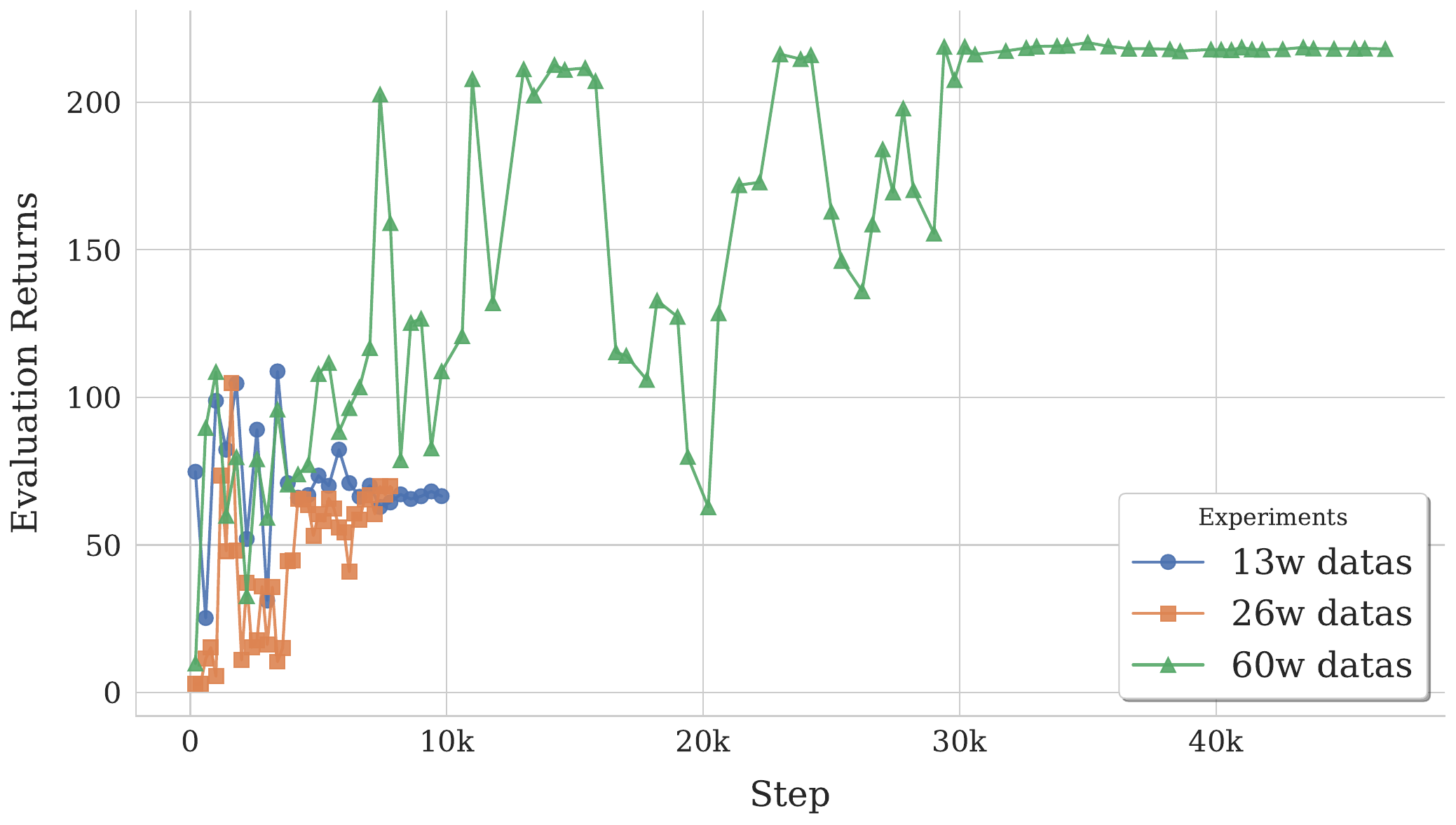}
        \caption{Returns 280}
        \label{fig:data_scaling_280}
    \end{subfigure}
    \caption{Maze2D experimental data scaling.}
    \label{fig:data_scaling}
\end{figure}

This section analyzes the scaling laws of our model with respect to both data volume and parameter count. The results are presented in Figure \ref{fig:data_scaling} and Table \ref{tab:maze2d_comparison}, respectively. In Table \ref{tab:maze2d_comparison}, we report the optimal performance for each model across all data scales, accounting for variations in the data volume required for convergence.

First, regarding data volume (Fig \ref{fig:data_scaling}), we observe a clear trend. When sampling from a fixed source dataset of 10 million steps, increasing the final training dataset size from 130k to 400k samples leads to monotonic performance improvements. This empirically validates the scaling law for data volume. Second, concerning model scale (Table \ref{tab:maze2d_comparison}), we see that performance generally improves as the parameter count increases from the DT-extended (i.e., with a parameter size of 720k) baseline up to 3B parameters. Additionally, we observed that the performance of the 3B-parameter model is on par with that of the 1.5B-parameter model. Our analysis indicates that the imitation learning-based supervised fine-tuning (SFT) paradigm has a performance ceiling, which in this task is already approached by the 1.5B model.

\subsubsection{Data Quality}
We evaluate the impact of data quality through two distinct sets of experiments. First, we quantify the performance gains attributable to our proposed data filtering methods. Second, we investigate the influence of the data collection policy by comparing models trained on data generated from a deterministic policy versus those trained on data from a stochastic, exploratory policy.

\paragraph{The impact of data filtering.}

\begin{figure}
    \centering
    \begin{subfigure}[b]{0.23\textwidth}
        \includegraphics[width=\textwidth]{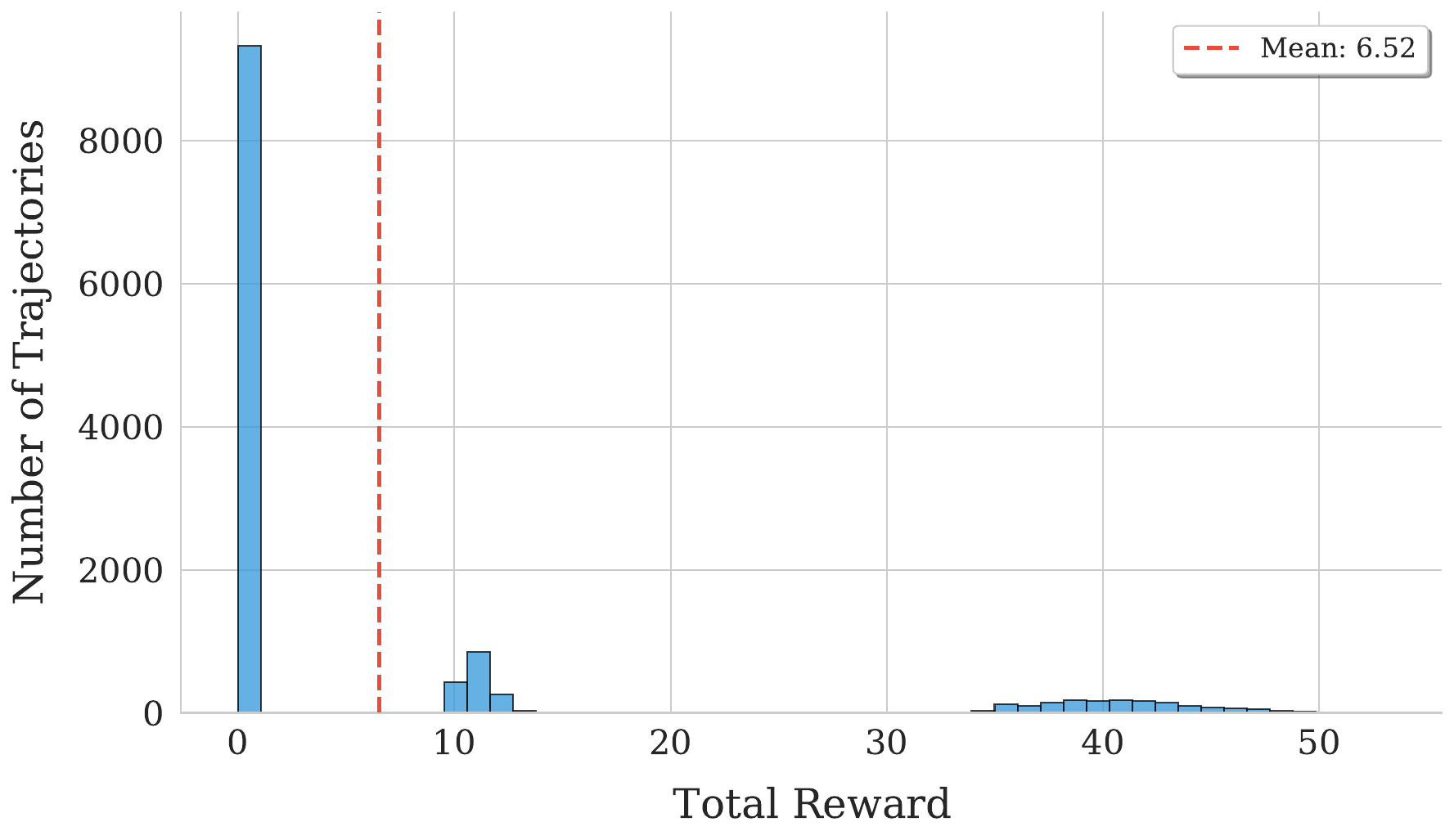}
        \caption{Reward distribution across all trajectories.}
        \label{fig:reward_dis_all_trajs}
    \end{subfigure}
    \hfill
    \begin{subfigure}[b]{0.23\textwidth}
        \includegraphics[width=\textwidth]{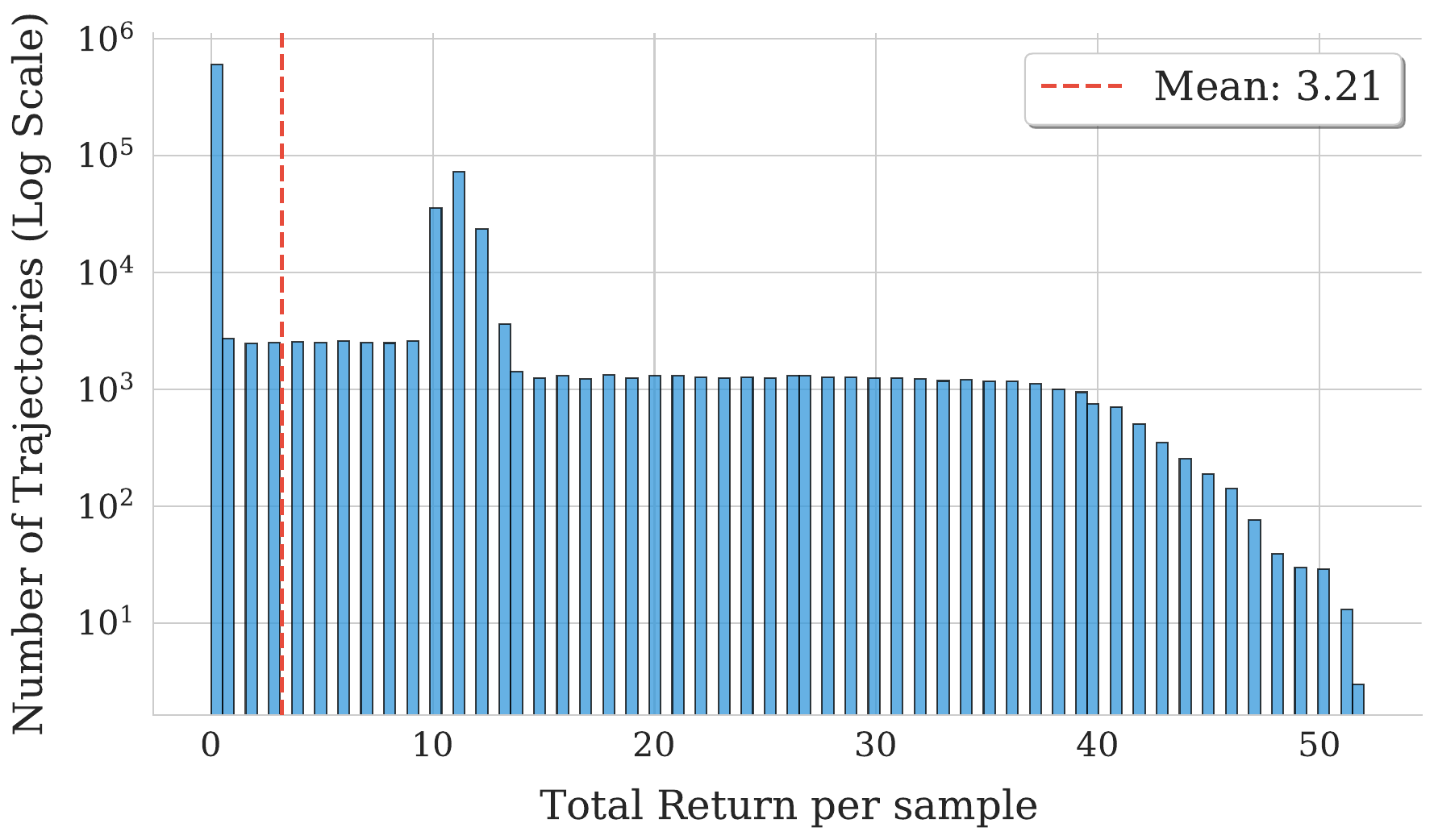}
        \caption{Initial returns distribution in sampled datas.}
        \label{fig:initial_returns_dis}
    \end{subfigure}
    \caption{Data distribution statistics.}
    \label{fig:data_dist_stat}
\end{figure}

The raw Maze2D offline dataset exhibits a severe long-tail distribution, with a preponderance of low-reward trajectories (Figure \ref{fig:reward_dis_all_trajs}). 
A naive shift-window approach (window size 20, yielding 800k samples) preserves this undesirable distribution, resulting in a training set dominated by low initial returns (Figure \ref{fig:initial_returns_dis}). 
Such data can hinder effective policy learning.

Therefore, we introduce a data pruning pipeline prior to subsequence sampling. 
Our method first removed all trajectories with a cumulative reward below the $\epsilon$ (set 0.5). 
From the remaining high-quality episodes, we then extract unique windowed subsequences of length 20. 
This curation process reduces the dataset from 800k raw samples to a focused set of 130k training examples, ensuring the model is primarily exposed to competent datas.

\begin{figure}
    \centering
    \begin{subfigure}[b]{0.23\textwidth}
        \includegraphics[width=\textwidth]{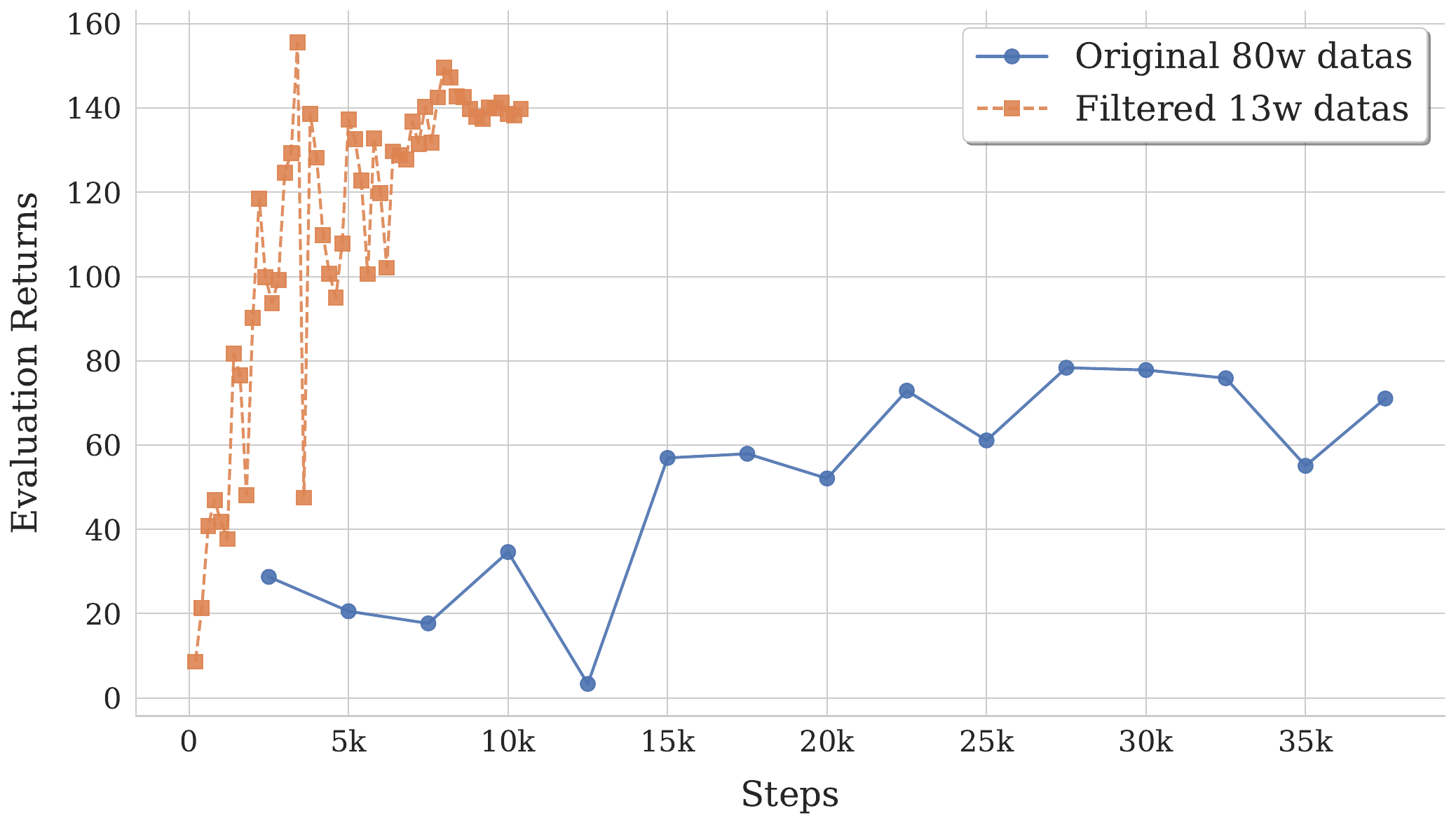}
        \caption{Returns 140}
        \label{fig:exp_wosample_140}
    \end{subfigure}
    \hfill
    \begin{subfigure}[b]{0.23\textwidth}
        \includegraphics[width=\textwidth]{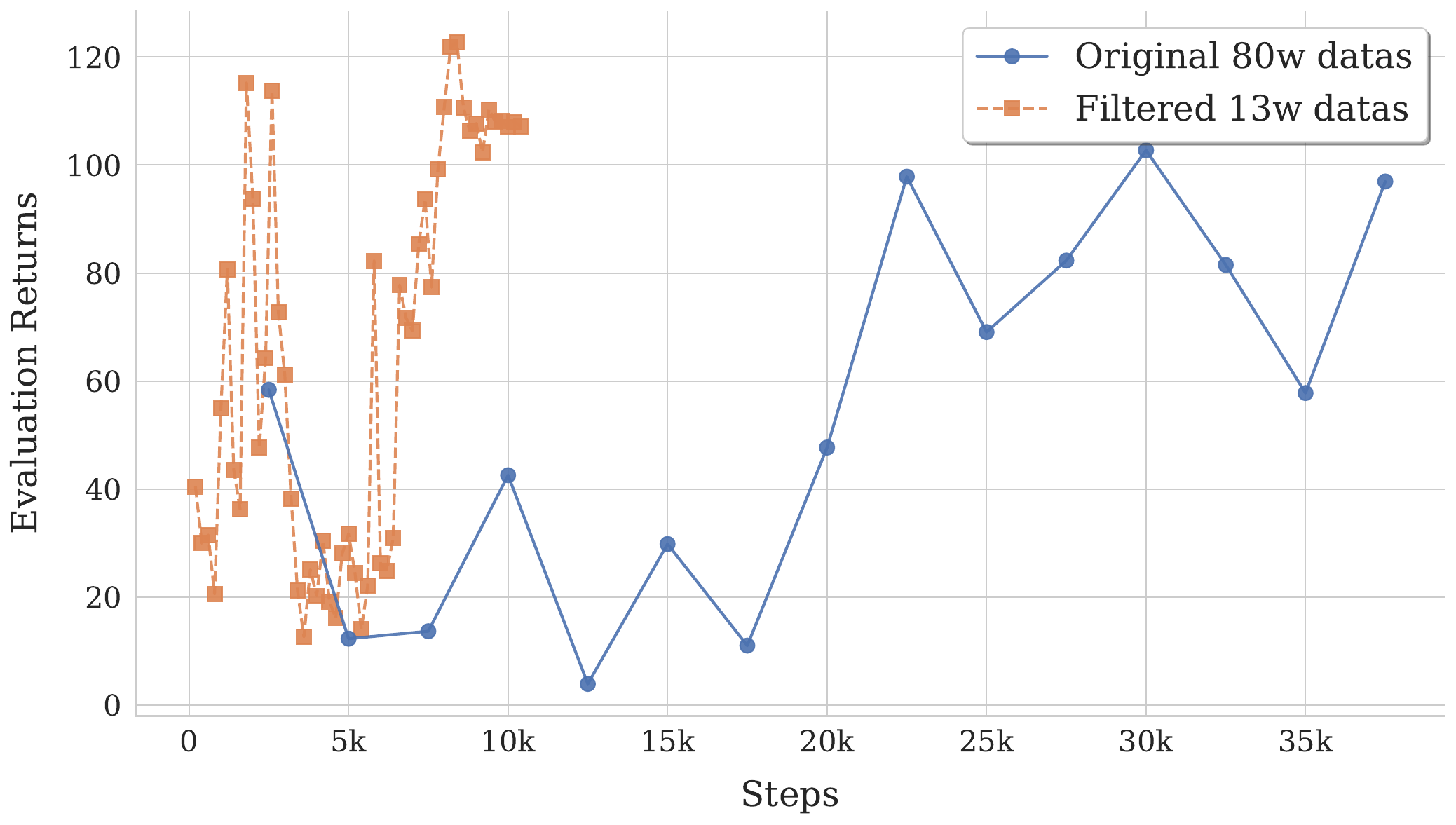}
        \caption{Returns 280}
        \label{fig:exp_wosample_280}
    \end{subfigure}
    \caption{Maze2D experimental performance w/o sample filtering}
    \label{fig:exp_wosample}
\end{figure}

On the one hand, the result of our data filtering method is clearly demonstrated in Figure \ref{fig:exp_wosample}. The filtered dataset enables the model to learn a more efficient policy, achieving target returns in significantly fewer steps. Furthermore, it substantially boosts the final performance, particularly for high target returns such as 140 and 280. These results underscore a crucial point: for imitation-based policy learning, data quality is a far more critical determinant of success than mere data quantity.

\begin{figure}
    \centering
    \begin{subfigure}[b]{0.23\textwidth}
        \includegraphics[width=\textwidth]{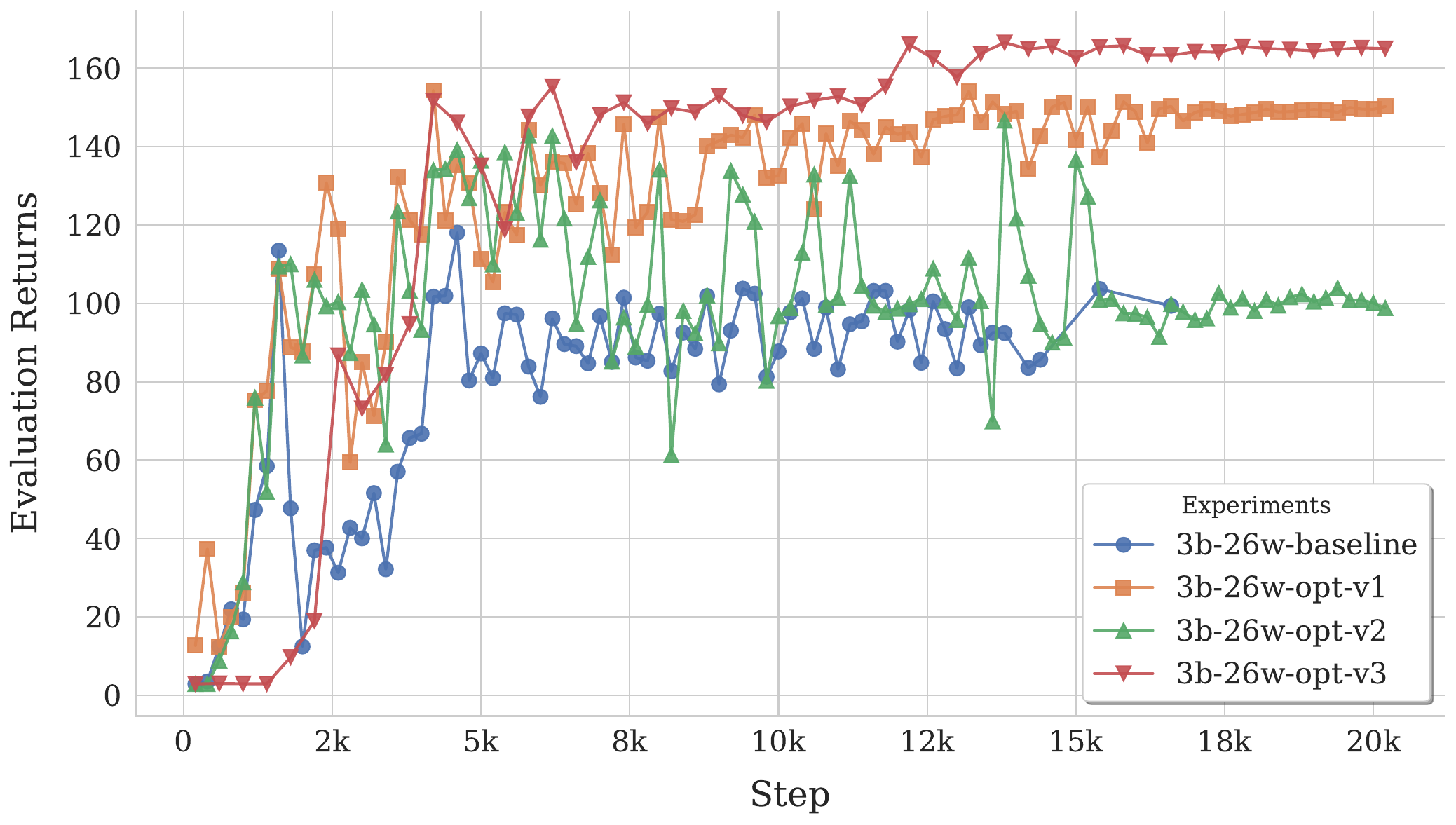}
        \caption{Returns 140}
        \label{fig:exp_lossopt_140}
    \end{subfigure}
    \hfill
    \begin{subfigure}[b]{0.23\textwidth}
        \includegraphics[width=\textwidth]{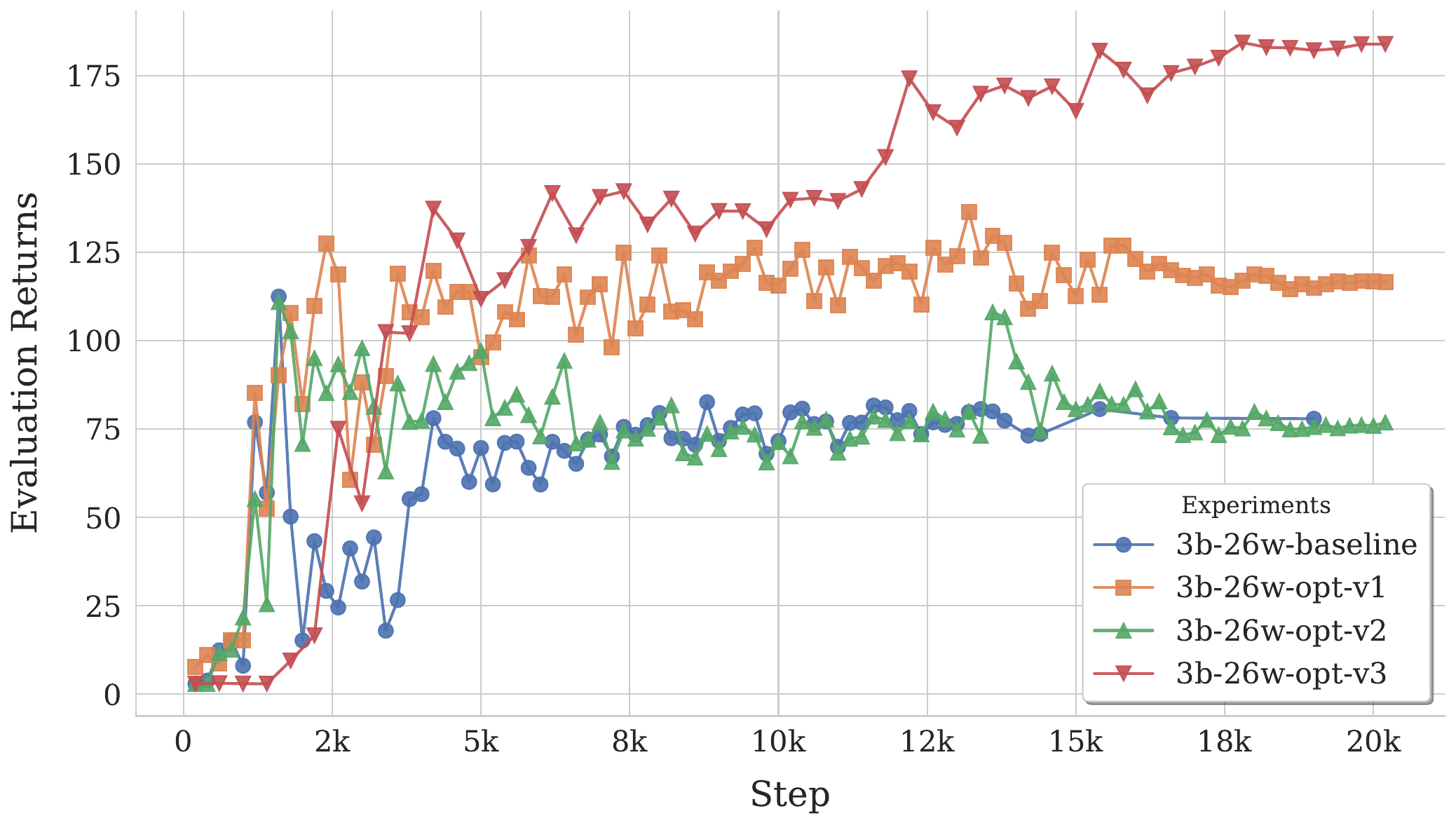}
        \caption{Returns 280}
        \label{fig:exp_lossopt_280}
    \end{subfigure}
    \caption{Maze2D experimental performance w/o sample filtering.}
    \label{fig:exp_lossopt}
\end{figure}

We further explore the actual effect of step-level filtering.
We evaluated three variants of loss optimization. loss-opt-v1 employs a hard filtering mechanism, masking out steps where rewards fall below a specific threshold. loss-opt-v2 adopts a softer approach, down-weighting these low-reward steps by a factor of 0.5 rather than discarding them. Finally, loss-opt-v3 extends the reweighting strategy of v2 by incorporating per-token normalization to balance the training objective.
A comparison of their performance is presented in Figure \ref{fig:exp_lossopt}. As illustrated in the figure, both v1 and v3 yield substantial benefits. They not only improve the stability of the training process but also enable the model to converge to a higher peak performance compared to the baseline. In comparison, while v2 performs similarly to the baseline, only marginally outperforming it, v3 delivers the most substantial performance improvement.

\paragraph{Impact of Exploration Diversity.}

\begin{figure}
    \centering
    \begin{subfigure}[b]{0.23\textwidth}
        \includegraphics[width=\textwidth]{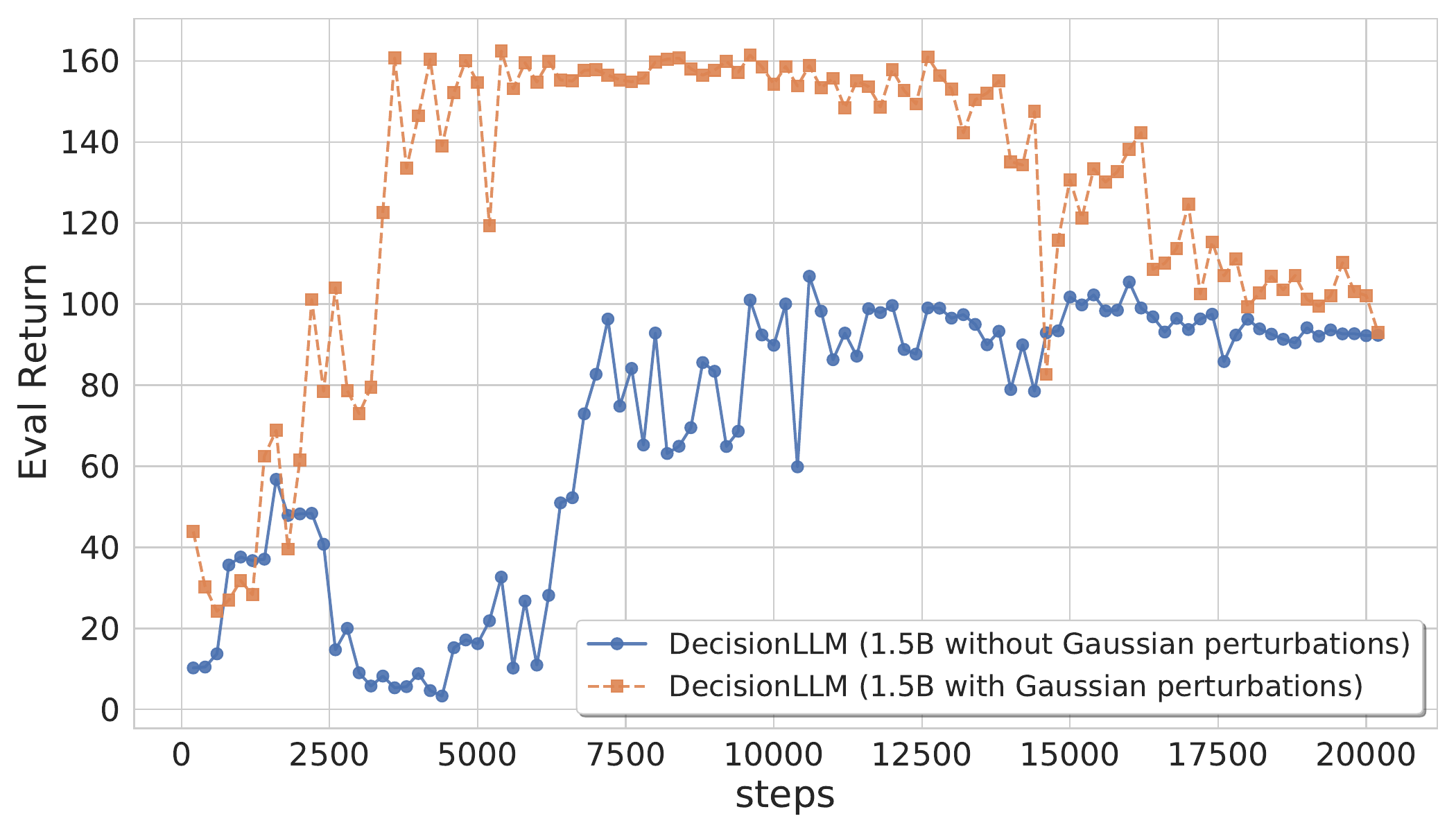}
        \caption{Reutrn 140}
        \label{fig:nois_nonoise_140}
    \end{subfigure}
    \hfill
    \begin{subfigure}[b]{0.23\textwidth}
        \includegraphics[width=\textwidth]{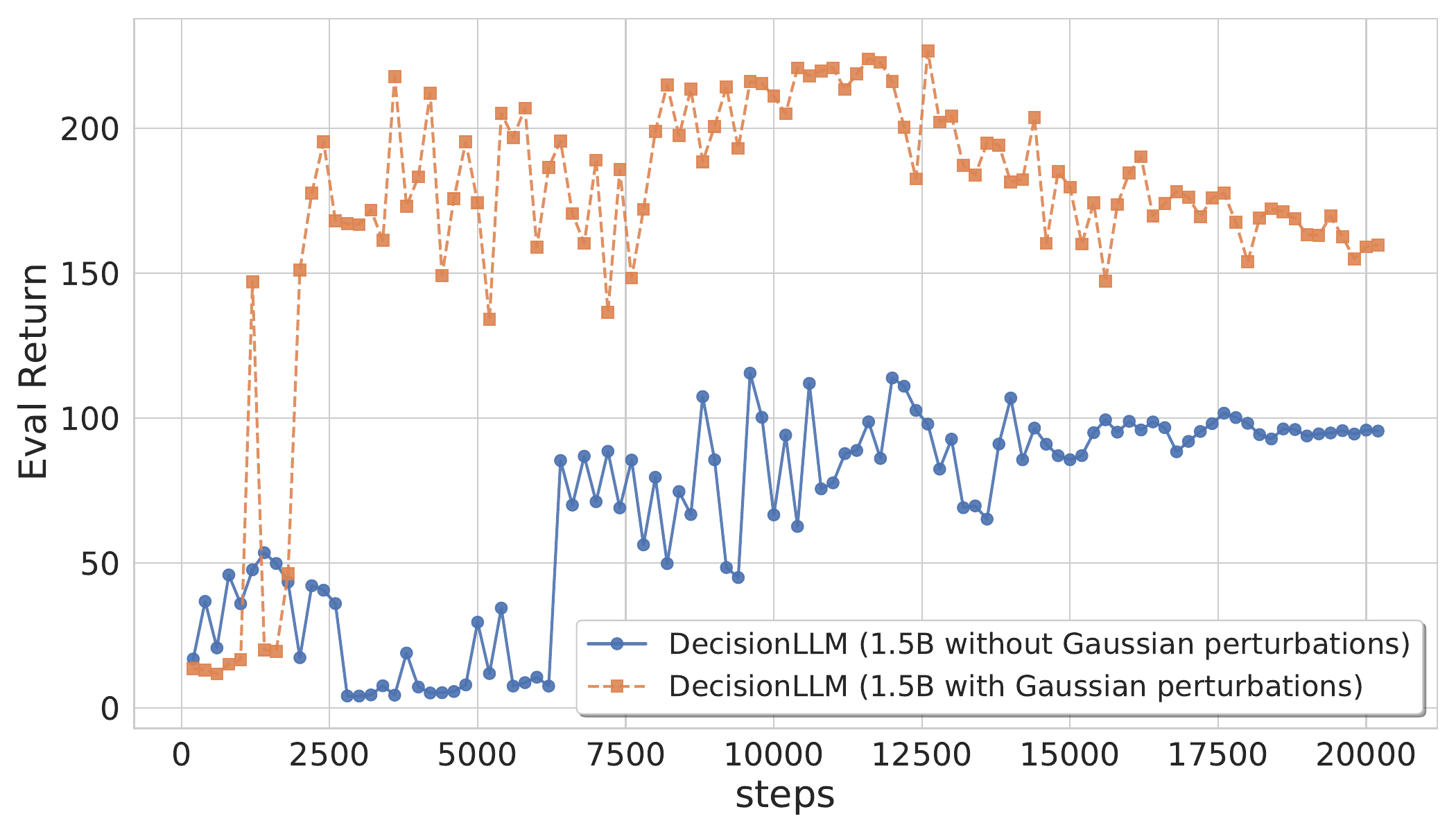}
        \caption{Return 280}
        \label{fig:nois_nonoise_140}
    \end{subfigure}
    \caption{Maze2D experimental performance with (without) noise in action policy.}
    \label{fig:w_o_noise_26w}
\end{figure}

To investigate the diversity of the sampling strategy, we conducted a simple experiment comparing perturbations during sampling with those without perturbations. Specifically, we sampled 10 million steps of data using the same strategy, and then used this strategy to sample 260k data samples with a window of 20. The only difference was whether Gaussian perturbations were added to the actual actions. We trained those two different dataset on Qwen2.5-1.5B-Instruct.The experimental results are shown in Figure \ref{fig:w_o_noise_26w}. The stochasticity of the data collection policy has a dramatic impact on final performance. Models trained on the exploratory dataset (with noise) reached a peak return of 220, a stark contrast to the 150 ceiling achieved with the deterministic action policy. This provides compelling evidence that a rich, exploratory training set is a key ingredient for training high-performing offline models. 



\subsubsection{Impact of Pretrained Parameters}

\begin{figure}
    \centering
    \begin{subfigure}[b]{0.23\textwidth}
        \includegraphics[width=\textwidth]{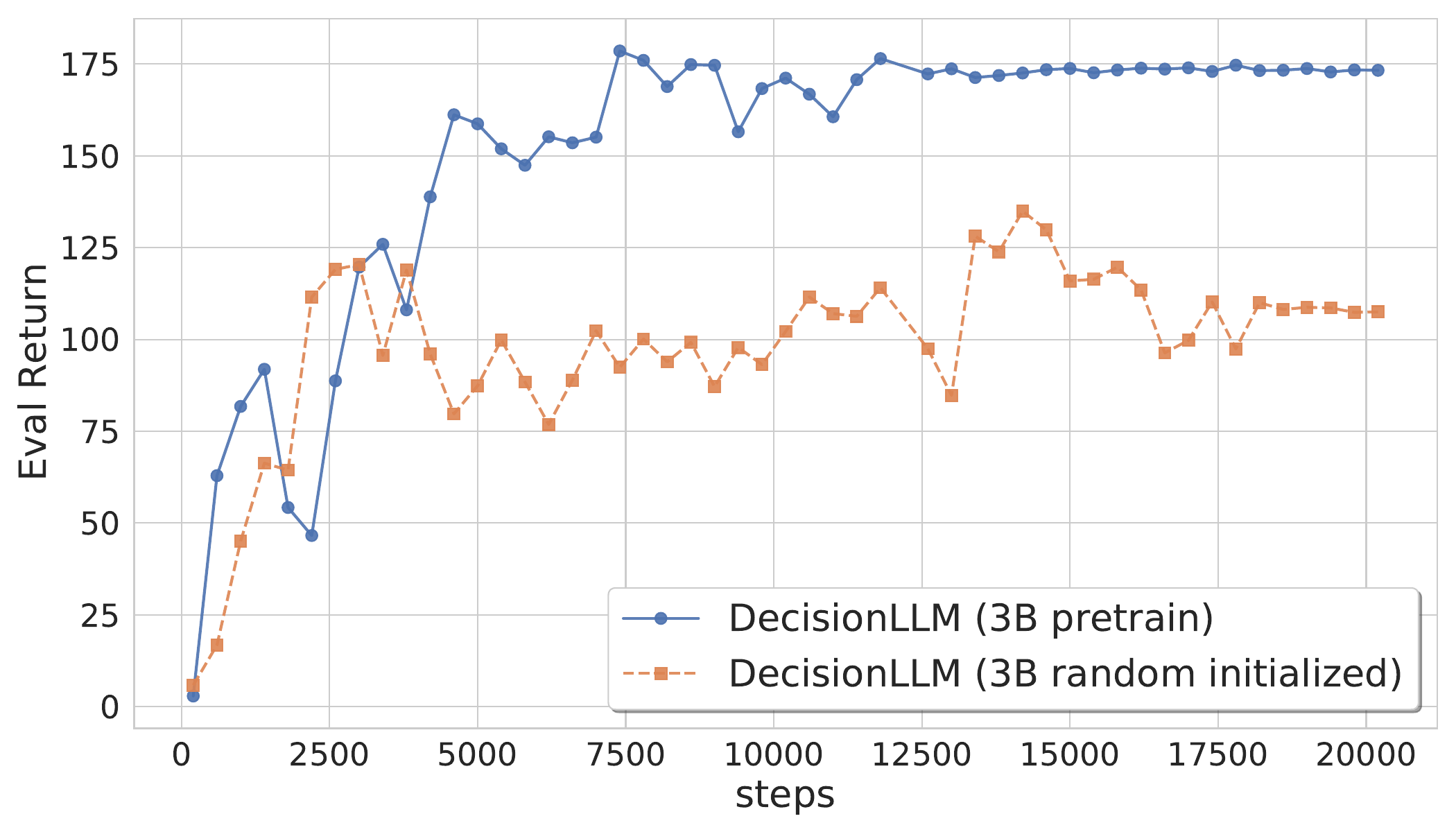}
        \caption{Returns 140}
        \label{fig:pretrain_140}
    \end{subfigure}
    \hfill
    \begin{subfigure}[b]{0.23\textwidth}
        \includegraphics[width=\textwidth]{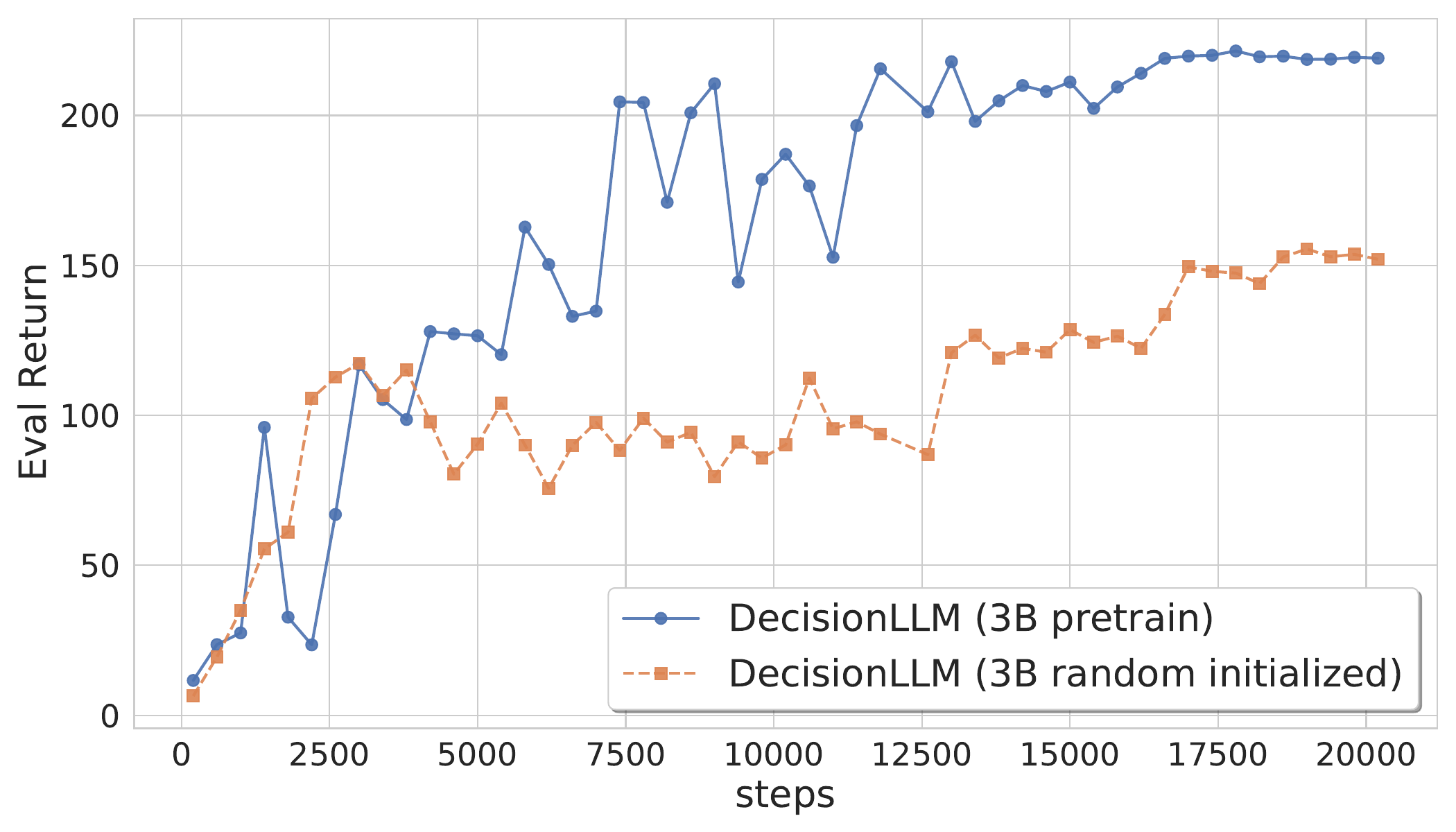}
        \caption{Returns 280}
        \label{fig:pretrain_280}
    \end{subfigure}
    \caption{Maze2D experimental performance with(without) pretrain.}
    \label{fig:w_o_pretrain}
\end{figure}

In this section, we conduct an ablation study to isolate the effect of the LLM's pretrained weights to final task performance. 
We compare two models under identical hyperparameter and data conditions, but one initialized with publicly available pretrained weights, and the other model trained with random initialization. 
As shown in Figure \ref{fig:w_o_pretrain}, the model initialized with pretrained parameters achieves higher returns and more stable convergence.

Our analysis suggests that even though LLM pretraining occurs exclusively on textual data, the foundational capabilities developed during this phase provide a strong inductive bias for decision making. 
Specifically, the model's highly developed sequence modeling and pattern recognition abilities, honed on vast text corpora, appear to transfer effectively, providing a superior starting point for learning the structure of trajectory prediction. More analysis can be found in Appendix \ref{appendix_pretrain_explanation}.


\section{Conclusions}
In this paper, we introduced DecisionLLM, a novel paradigm for LLMs to long sequence decision making. By treating trajectories as a distinct, non-textual modality, our approach successfully overcomes the inherent numerical insensitivity of LLMs and fully leverages sequential trajectory data. We demonstrated that by jointly modeling past trajectories and language instructions, DecisionLLM can effectively predict future actions in an autoregressive manner. Our systematic analysis revealed clear scaling laws with respect to both model size and data volume, providing valuable insights for future development. Furthermore, we presented a dual-level data curation methodology that significantly enhances performance by improving data quality. The empirical results on the challenging Maze2D and AuctionNet benchmark validate the superiority of our framework. Our flagship DecisionLLM-3B model achieves improvements of 69.4 and 0.085 over the traditional DT on the Maze2D umaze-v1 and AuctionNet task. These findings confirm that DecisionLLM represents a significant step forward in enabling LLMs to master complex, long-horizon control tasks.

\bibliography{sample-base}
\bibliographystyle{icml2026}

\newpage
\appendix
\onecolumn
\section{Performance Explanation}
\label{performance_appendix}

\begin{table}[htbp]
  \centering
  \caption{Comparison experiments (extension) on Maze2D-umaze-v1.}
  \label{tab:maze2d_comparison_ext}
  \begin{tabular}{lcc}
    \toprule
    Model & Return & Score \\
    \midrule
    DecisionLLM(0.5B)-base & $124.69\pm 110.74$ & $73.06 \pm 80.24$ \\
    DecisionLLM(1.5B)-base & \textbf{$186.37 \pm 87.74$} & \textbf{$117.76 \pm 63.58$} \\
    DecisionLLM(3B)-base & $ 153.29 \pm 50.30$ & $93.79 \pm 36.44$ \\
    \bottomrule
  \end{tabular}
\end{table}

Table \ref{tab:maze2d_comparison_ext} presents our base DecisionLLM, trained on a 130k-sample dataset curated from the original 1M D4RL steps via trajectory filtering and window sampling. Even on this compact dataset, our model significantly outperforms the DT baseline and exhibits robust scaling from 0.5B to 1.5B parameters. However, the 3B model's performance on this base dataset was unexpectedly poor, which we attributed to under-convergence. We tested this theory by expanding the dataset to train our flagship models (Table \ref{tab:maze2d_comparison}). The resulting surge in the 3B model's performance provides conclusive evidence for our hypothesis, demonstrating that the full capacity of large-scale DecisionLLM models is unlocked only when matched with sufficient data.

\section{Related Works}

\subsection{Long Sequence Decision Making}
Recently, long-sequence decision-making problems have been modeled as a Markov Decision Process (MDP), which assumes the Markov state transition. A critical context for this problem is the offline scenario, where the training data is obtained entirely through pre-trained offline sampling. Previous approaches have primarily relied on offline RL methods, which primarily mitigate the impact of distributional shift \cite{fujimoto2019off,kidambi2020morel,siegel2020keep} or learn the generalization ability of the model through offline datasets \cite{ajay2020opal,singh2020parrot,eysenbach2018diversity,lu2020reset}. After that, a prominent generative paradigm reframes this problem by modeling the probability distribution of future actions conditioned on historical trajectories. In such methods, they fits the probability distribution of future actions and historical trajectories, which can be further empowered based on the powerful capabilities of Transformer \cite{vaswani2017attention, chen2021decision} and Diffusion \cite{rombach2022high, chi2023diffusion, wang2022diffusion, guo2024generative}. Two classes of models have been central to this shift: Transformer \cite{vaswani2017attention, chen2021decision} and Diffusion \cite{rombach2022high, chi2023diffusion, wang2022diffusion, guo2024generative}. The success of the two frameworks depends on the strong basic capabilities of the model, so it is worth exploring whether LLM can bring better improvements to long-sequence decision-making problem.

\subsection{LLM for Long Sequence Decision}
Large language models have shown good results in similar sequential decision-making such as autonomous driving \cite{li2025recogdrive, cui2023drivellm} and robotics \cite{kim2024openvla, intelligence2504pi0}. We noticed that semantic space plays a more obvious role in the process, which usually requires accurate semantic description of clear states, actions, and goals in each scenario. While powerful, this semantic representation is a product of the textual modality. Its structure is therefore inherently discrete and symbolic, lacking the native capacity to represent the continuous, high-dimensional vector spaces that characterize the dynamics of most sequential decision-making environments \cite{dziri2023faith}. 

Consequently, many contemporary LLM-based agents rely on modular designs, integrating separate reinforcement learning (RL) components to handle tasks such as reward shaping \cite{qu2025latent}, exploration \cite{hao2025llm}, or data augmentation \cite{pang2024kalm, wan2025think}. However, these hybrid methods have inherent limitations and often lack generalizability across all scenarios. In contrast, our work explores the potential of using LLMs as direct, end-to-end decision-makers. This direction has been largely unexplored, partly because the performance of earlier approaches was constrained by the limited capacity of their underlying models \cite{pang2024kalm}. Therefore, investigating whether today's powerful, large-scale LLMs can directly and effectively solve these tasks is a research question of significant practical importance.

\section{Maze2D Prompt}
The specific prompt template used in our experiments is provided as follows:
\begin{tcolorbox}[
    breakable,  
    enhanced,
    title=Maze2D prompt,
    fonttitle=\bfseries,
    colback=white,
    colframe=blue!50!black,
    colbacktitle=blue!50!white,
    coltitle=black,
    boxrule=1pt,
    arc=3mm,
    left=6pt,
    right=6pt,
    top=8pt,
    bottom=8pt,
    attach boxed title to top left={xshift=4mm,yshift=-2mm},
    boxed title style={colframe=blue!50!black, colback=blue!50!white, sharp corners}
]
\texttt{You are a maze navigation expert. Your goal is to reach the destination from your current position using the fewest steps possible. You receive a reward of +1 for reaching the destination; all other positions have a reward of 0. You need to choose the optimal movement to maximize the total reward. Each state at every time step is represented by four values [x, y, vx, vy]:}

\begin{itemize}
    \item \texttt{(x, y) represents the current position coordinates}
    \item \texttt{(vx, vy) represents the current velocity}
    \item \texttt{All values range from [-1.0, 1.0].}
\end{itemize}

\texttt{The action at each time step is a 2D vector: [ax, ay]}

\begin{itemize}
    \item \texttt{ax represents the control force (acceleration) applied in the x-axis direction}
    \item \texttt{ay represents the control force applied in the y-axis direction}
    \item \texttt{All values range from [-1.0, 1.0].}
\end{itemize}

\texttt{Each step has a corresponding "Returns-to-Go" value, a scalar representing the expected cumulative reward from the current time step to the end of the trajectory.}

\texttt{You will receive trajectory information, including the state sequence, action sequence, and Returns-to-Go sequence for a complete episode, formatted as follows:<|traj\_begin|><|traj\_end|>}

\texttt{Your task is to learn a policy based on this trajectory data: given the current state and its corresponding Returns-to-Go, predict the optimal action a to take at that time step.}

\texttt{Please explain your understanding of the current policy and output the corresponding action value, along with an explanation.}
\end{tcolorbox}

\section{The Explanation of Pretrained Parameters}
\label{appendix_pretrain_explanation}

\begin{figure}
    \centering
    \begin{subfigure}[b]{0.35\textwidth}
        \includegraphics[width=\textwidth]{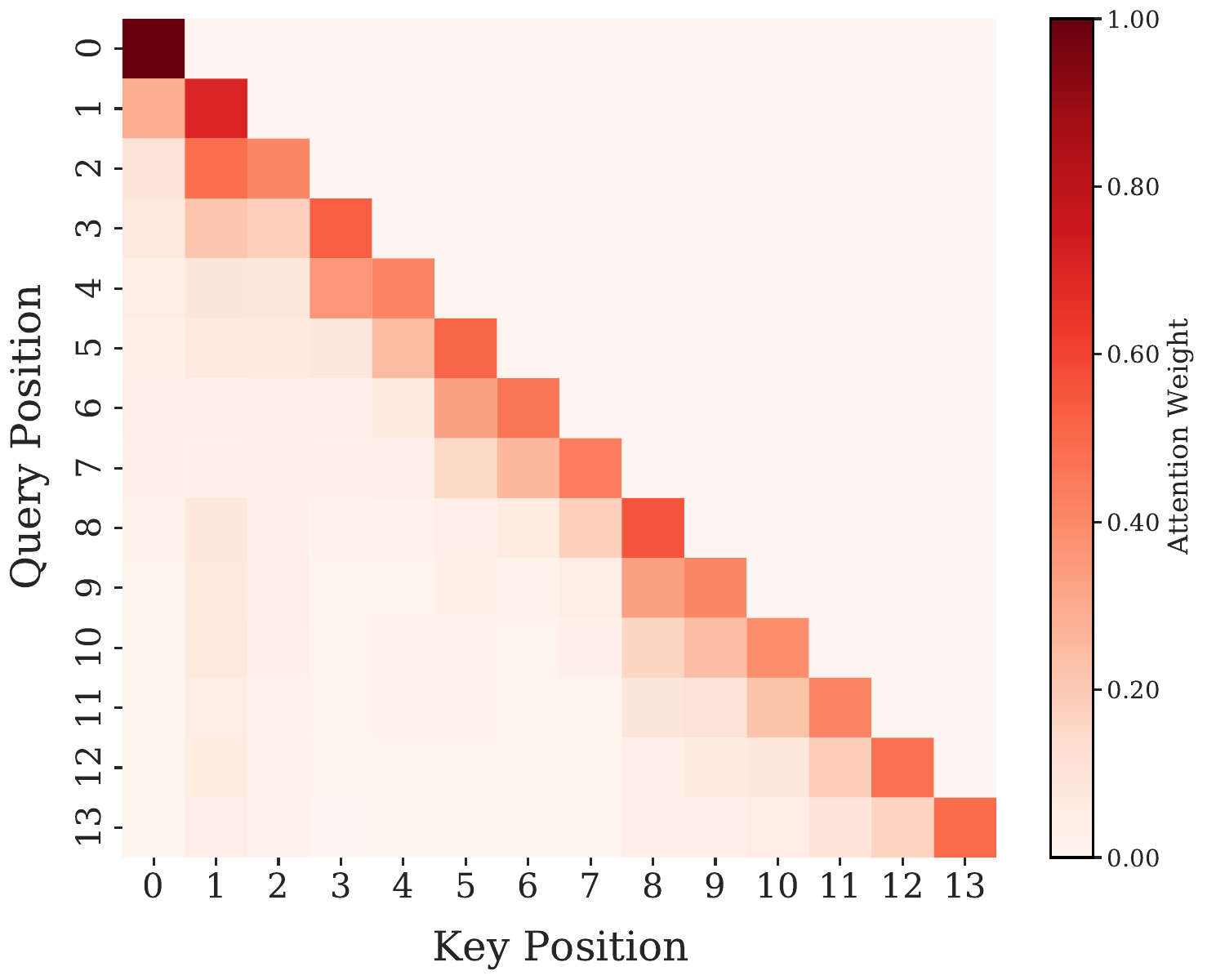}
        \caption{Pretrained Model}
        \label{fig:pretrain_attention_map}
    \end{subfigure}
    \hspace{0.1\textwidth} 
    \begin{subfigure}[b]{0.35\textwidth}
        \includegraphics[width=\textwidth]{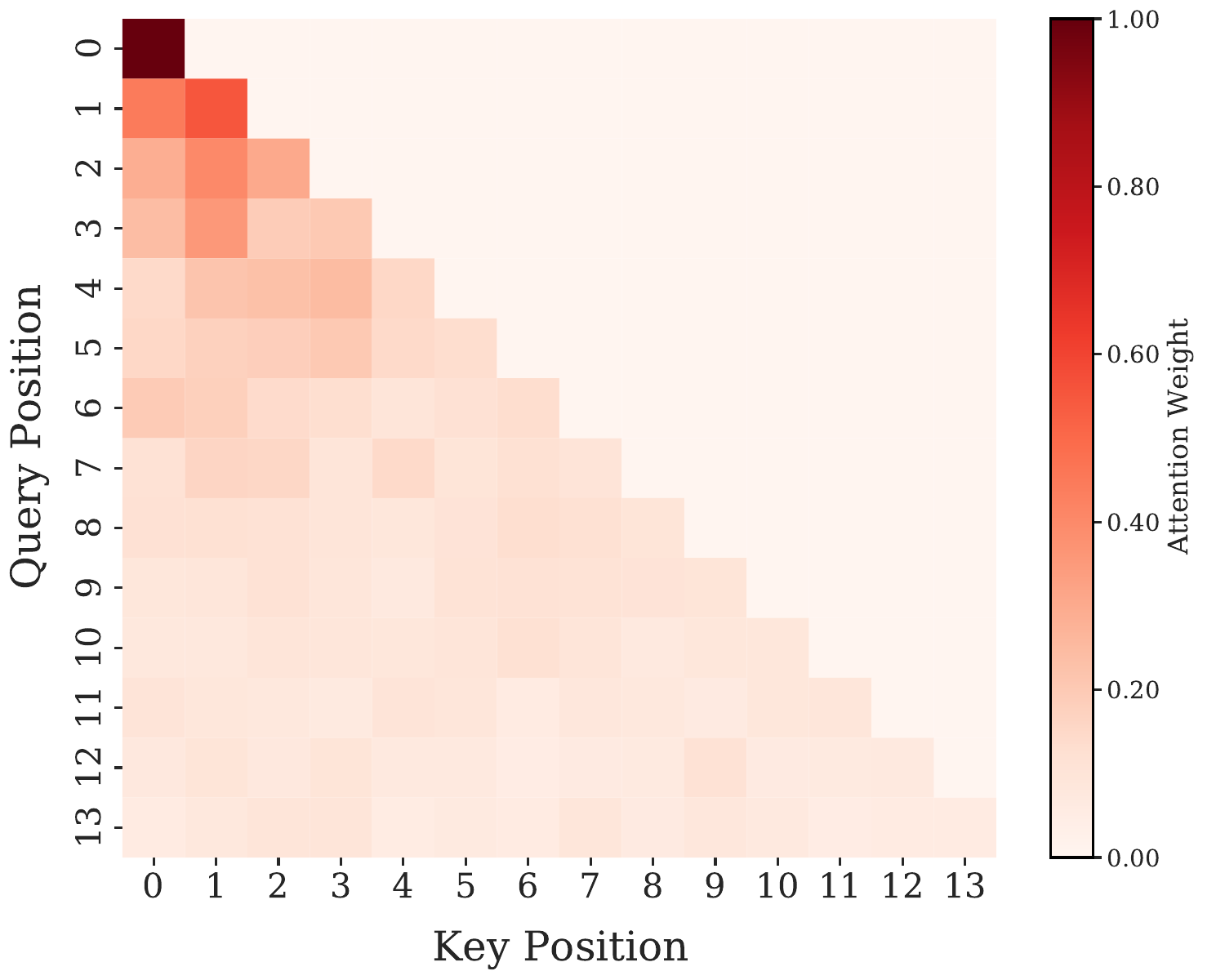}
        \caption{Random Initialized}
        \label{fig:random_attention_map}
    \end{subfigure}
    \caption{Attention distribution map.}
    \label{fig:attention_dis}
\end{figure}

To explain the impact of pretrained parameters on model behavior, we conducted an analysis of their intrinsic attention patterns. We probed the models' responses to a semantically null input string (e.g., \texttt{z\$x-$\alpha\beta$hwoqa\%\^{}\&*()<>?:"}), which simulates an encounter with an incomprehensible sequence and isolates learned structural biases. As illustrated in Figure \ref{fig:attention_dis}, the attention matrix of the pretrained model, averaged across all heads in the first layer, exhibits a highly structured pattern. Its attention is predominantly concentrated on the last token, with a sparse but deliberate allocation to preceding tokens.

In stark contrast, the randomly initialized model displays a diffuse and unstructured attention distribution. This comparison reveals that pretraining endows the model with a crucial inductive bias: a strong focus on recent information. In the context of long-sequence decision-making, where not all historical steps are equally relevant to future rewards, this learned "recency bias" is highly advantageous. It allows the model to efficiently prioritize the most recent actions while attending to relevant past context. Consequently, this superior initialization facilitates more efficient convergence and enables a higher ultimate performance ceiling.

\section{Sensitivity Analysis of Rtgs at Inference}
\begin{table}[htbp]
  \centering 
  \caption{Model sensitivity to the initial target return.}
  \label{tab:sensitivity} 
  \begin{tabular}{ccc}
    \toprule
    Initial Rtgs & Predicted Rtgs & Score \\
    \midrule
    100 & $152.04\pm27.90$ & $92.88\pm20.22$\\
    120 & $162.66\pm30.34$ & $100.58\pm21.99$ \\
    140 & $173.28\pm29.14$ & $108.27\pm21.12$\\
    160 & $180.51\pm24.71$ & $113.51\pm17.91$ \\
    180 & $190.29\pm26.98$ & $120.60\pm19.55$ \\
    200 & $202.31\pm33.53$ & $129.31\pm24.30$ \\
    220 & $207.56\pm36.98$ & $133.11\pm26.79$ \\
    240 & $209.33\pm38.09$ & $134.40\pm27.60$ \\
    260 & $214.47\pm43.25$ & $138.12\pm31.34$ \\
    280 & \textbf{$219.08\pm51.36$} & \textbf{$141.46\pm37.21$} \\
    300 & $217.11\pm 54.55$& $140.03\pm 39.53$ \\
    \bottomrule
  \end{tabular}
\end{table}
We further evaluate the model's sensitivity to the initial target Rtgs. Specifically, during inference, we set the initial Rtgs from 100 to 280, with an interval of 20. The experimental results are shown in Table \ref{tab:sensitivity}. We can see that when the initial Rtgs ranges from 100 to 280, the actual Rtgs predicted by the model exhibit good monotonicity. However, after 280, the model's Rtgs begins to decline. This is because 300 is the theoretical upper limit for this task, meaning the model cannot effectively fit Rtgs outside the learnable range. Therefore, it is necessary to set a valid Rtgs during inference.

\section{Experimental hyperparameters and data description}
\label{exp_appendix}

\subsection{Comparison Experiments Setting (Table \ref{tab:maze2d_comparison})}
\label{appendix_comparsion}
For the comparison experiments, we used a standardized set of hyperparameters. The data was sampled from 10 million trajectories. For models with different numbers of parameters, we performed data scaling experiments until a converged result was obtained, which we then used as our final result. For data optimization, we uniformly used the loss-opt-v3 optimization method with trajectory filtering and step filtering, setting the hyperparameters $\epsilon$ and $\beta$ to 0.5. All other training parameters were identical to those mentioned previously.

\subsection{Scaling Laws Setting}
The hyperparameter configurations, data sources, and specific sampling sizes for Figure \ref{fig:data_scaling} and Table \ref{tab:model_scaling} are detailed in Appendix \ref{appendix_comparsion} and Section 4.3, respectively. Therefore, we refer the reader to these sections for complete details.

\subsection{Data Quality Setting}
Figure \ref{fig:exp_wosample} presents the results of the ablation study on our trajectory-level filter. This experiment was conducted using an initial dataset of 1M steps from D4RL datasets, with the specific sampling methodology and resulting data counts detailed in Section 4.4. Conversely, Figure \ref{fig:exp_lossopt} illustrates the ablation study for the step-level filter. For this analysis, 260k training datas with window size 20 were sampled from an expanded dataset of 10 million steps. All other hyperparameters were held consistent with the previously described experimental setup.

\subsection{Impact of Pretrained Parameters's Setting}
\label{appendix_parameter}
Figure \ref{fig:w_o_pretrain} presents the training results for our 3B parameter model. The model was trained on a dataset of 130k windowed samples, which were curated from an expanded data pool of 10 million trajectories. For this specific experiment, a minor hyperparameter adjustment was made: the batch size was set to 32. This modification was implemented to enhance training stability on our computational cluster.

\subsection{Sensitivity Analysis's  Setting}
The 3B parameter model presented in Table \ref{tab:sensitivity} was trained on a dataset of 130k samples, curated from an augmented pool of 10 million trajectories. During training, a batch size of 32 was used, consistent with the configuration detailed in the Appendix \ref{appendix_parameter}. Notably, all reported results are based on the model checkpoint from the final training step, rather than a checkpoint selected for peak performance on a validation set.

\end{document}